\documentclass[journal,twoside,web]{ieeecolor}
\usepackage{tmi}
\usepackage{cite}
\usepackage{amsmath,amssymb,amsfonts}
\usepackage{algorithmic}
\usepackage{graphicx}
\usepackage{textcomp}

\usepackage{booktabs}
\usepackage{multirow}

\usepackage[colorlinks]{hyperref}
\def\BibTeX{{\rm B\kern-.05em{\sc i\kern-.025em b}\kern-.08em
    T\kern-.1667em\lower.7ex\hbox{E}\kern-.125emX}}
\begin{document}
\title{Evidential Calibrated Uncertainty-Guided Interactive Segmentation paradigm for Ultrasound Images}
\author{Jiang Shang, Yuanmeng Wu, Xiaoxiang Han, Xi Chen and Qi Zhang
\thanks{This work was supported by the National Natural Science Foundation of China (Grant No. 62071285), the Eastern Scholars Program from Shanghai Municipal Education Commission, and the Shanghai Technical Service Center of Science and Engineering Computing, Shanghai University. }
\thanks{Jiang Shang, Xiaoxiang Han and Qi Zhang are with the SMART (Smart Medicine and AI-based Radiology Technology) Lab, Shanghai Institute for Advanced Communication and Data Science, School of Communication and Information Engineering, Shanghai University, Shanghai, 200444, China (e-mail: jiangshang@shu.edu.cn; hanxx@shu.edu.cn; zhangq@t.shu.edu.cn).}
\thanks{Yuanmeng Wu is with the Software Engineering Institute, East China Normal University, Shanghai, 200062, China (e-mail: 71265902106@stu.ecnu.edu.cn).}
\thanks{Xi Chen is with the College of liberal arts and science, University of Illinois Urbana-Champaign, Urbana, IL, USA (e-mail: chenxi0107@gmail.com ).}
\thanks{Jiang Shang and Yuanmeng Wu contributed equally to this
	work. Corresponding author: Qi Zhang}}
\maketitle

\begin{abstract}
Accurate and robust ultrasound image segmentation is critical for computer-aided diagnostic systems.
% , as accurate and robust segmentation results contribute to disease assessment and treatment.
Nevertheless, the inherent challenges of ultrasound imaging, such as blurry boundaries and speckle noise, often cause traditional segmentation methods to struggle with performance.
Despite recent advancements in universal image segmentation, such as the Segment Anything Model, existing interactive segmentation methods still suffer from inefficiency and lack of specialization. These methods rely heavily on extensive accurate manual or random sampling prompts for interaction, necessitating numerous prompts and iterations to reach satisfactory performance.
In response to this challenge, we propose the Evidential Uncertainty-Guided Interactive Segmentation (EUGIS), 
an end-to-end, efficient tiered interactive segmentation paradigm based on evidential uncertainty estimation for ultrasound image segmentation. 
Specifically, EUGIS harnesses evidence-based uncertainty estimation, grounded in Dempster-Shafer theory and Subjective Logic, to gauge the level of uncertainty in the predictions of model for different regions. 
By prioritizing sampling the high-uncertainty region, our method can effectively simulate the interactive behavior of well-trained radiologists, enhancing the targeted of sampling while reducing the number of prompts and iterations required.
Additionally, we propose a trainable calibration mechanism for uncertainty estimation, which can further optimize the boundary between certainty and uncertainty, thereby enhancing the confidence of uncertainty estimation. 
Extensive experiments on three ultrasound datasets demonstrate the competitiveness of EUGIS against the state-of-the-art non-interactive segmentation and interactive segmentation models.
We believe this new paradigm will provide a novel perspective for the field of interactive segmentation and is expected to promote further development of interactive image segmentation for ultrasound image. 
Code and data will be available at \textcolor{blue}{\href{}{https://github.com/JiangVentinal/EUGIS}}.
\end{abstract}


\section{Introduction}
Medical image segmentation, the process of accurately identifying and delineating specific tissues or lesions within medical images, plays a pivotal role in computer-aided diagnosis systems\cite{unet, unet++, csvt2, csvt4}.
Propelled by rapid advancements in deep learning and bolstered by the non-invasive nature and convenience of ultrasound imaging, end-to-end neural networks (NNS) have demonstrated significant potential in the domain of automated segmentation for ultrasound modalities\cite{aaunet, h2former, cmunet, csvt1, csvt5}.
Nonetheless, due to the inherently complex scenarios present in ultrasound images\cite{bianjie1, noise1}, such as blurred edges and speckle noise, there remains a significant challenge to the attainment of stable segmentation outcomes and the development of deep learning models endowed with robust generalization strengths.

In the realm of interactive segmentation, universal foundation models for vision segmentation such as Segment Anything Model (SAM) have demonstrated formidable robust zero-shot generalization capacity, effectively mitigating the limitations of traditional neural networks in terms of generalization performance. 
The strength of such methods lies in their ability to allow users to guide the model in refining segmentation results by continuously providing simple prompts, such as points, not only making the segmentation of target objects easier but also enabling the model to adapt to various medical image segmentation tasks.
While extensive user interactions typically results in more robust segmentations, it is crucial that an effective interactive segmentation approach should generate high-quality segmentation results with minimal interaction prompts.
This implies that it ought to realize more accurate segmentation with fewer prompts (points) and iterations, thereby diminishing the user’s burden and augmenting efficiency.

\begin{figure}[t]
	\centering
	\includegraphics[width=\linewidth]{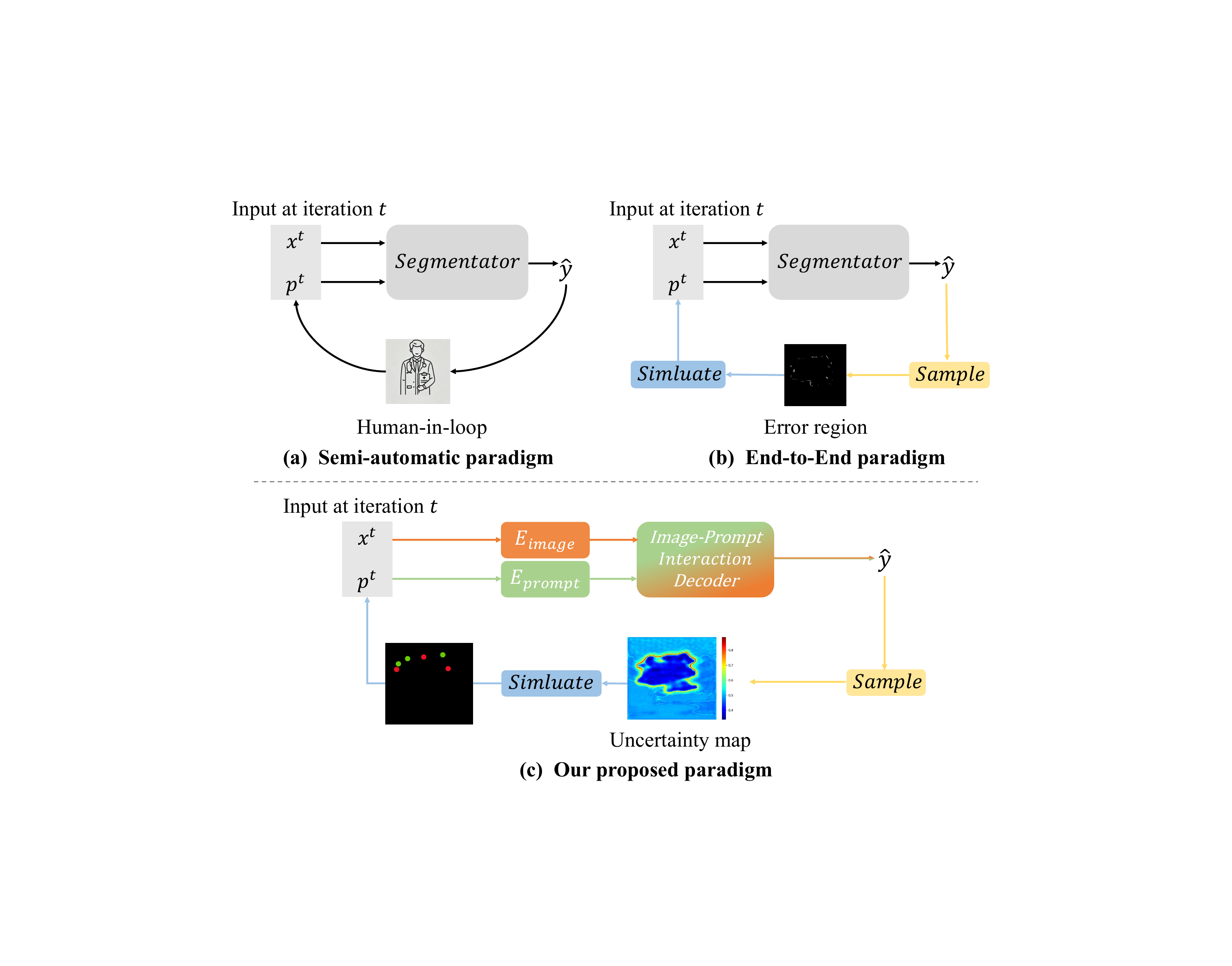}
	\caption{Different interactive segmentation paradigms, $x^t$ and $p^t$ refer to the input image and prompt in the iteration $t$, respectively. (a) Semi-automatic paradigm with the human-in-loop interaction process; (b) End-to-End paradigm with human-simulated prompt information; (c) Our proposed method with prompts generated by uncertainty map.}
	\label{fig:lk}
\end{figure}

Thus far, a series of studies in interactive segmentation have been proposed for medical image.
These works can be generally grouped into two categories:
(1) Semi-automatic interactive segmentation involves human interaction in the loop\cite{banzidong1, banzidong3, banzidong2}; (2) End-to-end interactive segmentation with automated simulation of human interaction\cite{masam, sammed2d, samu, desam, medsam, msa, samed}. 
In the first category, semi-automatic interactive segmentation approaches typically involve interactions  provided by clinicians to generate more specialized prompts. 
This strategy leverages the domain knowledge and expertise of radiologists, making the prompts more targeted.
It significantly mitigates the inherent complexity of medical images and addresses the challenges posed by the diversity of various diseases or lesions,  while also efficiently guiding the model in segmentation and optimization. 
However, semi-automatic methods are not end-to-end training approaches, which may limit the system’s consistency and performance, and increase the complexity of the workflow. 
Furthermore, this approach heavily relies on the prompts from clinicians, which can lead to a dependency of the model on the clinicians’ expertise.
The second category involves studies that aim to implement end-to-end interactive segmentation procedures. 
These methods automatically simulate prompts based on the predicted segmentation and ground truth segmentation result, without human-in-the-loop providing the prompts.
In simple terms, these methods employ random sampling strategies to automatically generate prompts by sampling points in regions where the model’s predictions exhibit inaccuracies, thereby simulating human interactive behavior.
Despite the capability of aforementioned approaches in facilitating an end-to-end training process, the quality of the generated prompts and the simulation outcome remain suboptimal.
For instance, the random sampling strategy lacks specificity, which may fail to accurately locate the region where the model requires the most improvement, thus not effectively guiding the learning of the model.  
Additionally, due to the randomness of the prompts, the model may necessitate more prompts and iterations to attain optimal performance.
Thus, we explore the following question:
Could we formulate a sampling strategy that closely emulates the interactive behavior of expert clinicians, efficiently providing the model with the necessary prompts, thereby enabling the generation of higher-quality segmentation results with fewer interactive prompts while maintaining an end-to-end training process?

Inspired by the properties of Dirichlet-based evidential deep learning model\cite{evidentialdeeplearining, dirichletuncertainty}, we propose an elegant, robust, click-based approach for end-to-end interactive segmentation in ultrasound imaging, called \textbf{E}vidential \textbf{U}ncertainty-\textbf{G}uided Tiered \textbf{I}nteractive \textbf{S}egmentation (EUGIS).
Through the utilization of evidential uncertainty to offer efficient confidence assessment to closely simulate the interactive behavior of expert clinicians, our approach prioritizes providing the model with point prompts in regions of high confidence of uncertainty, which enables superior performance with fewer prompts and iterations, and making it possible to flexibly handle various downstream segmentation tasks. 
Furthermore, we calibrate the uncertainty estimation by following the guideline that the model should exhibit certainty when making accurate predictions and generate high uncertainty estimates when making poor predictions, in order to maximize the clarity of the boundary between certainty and uncertainty.
Additionally, we use hybrid model to comprehensively capture both local and global information, and employ multiple segmentation heads to generate multiple segmentation results and confidence scores, selecting the segmentation result with the highest confidence score to optimize predictions and enhance the robustness of the model.

Our proposed EUGIS was evaluated using three datasets with different tasks, namely breast, thyroid, and left ventricle. 
The results demonstrate that EUGIS with only a single point prompt outperforms state-of-the-art (SOTA) non-interactive and interactive segmentation methods, surpassing all other approaches.
This also indicates that developing evidential uncertainty-guided interactive segmentation method is promising as a novel end-to-end interactive segmentation paradigm.

Our main contributions can be summarized as:
\begin{itemize}
	\item[$\bullet$] We propose EUGIS, a novel interactive segmentation paradigm that simulates the interaction behavior of expert radiologists through point prompts generated by evidence-based uncertainty estimation, achieving high-quality segmentation results and generalization performance with fewer interactions.
	
	\item[$\bullet$] We introduce evidential uncertainty via Subjective Logic (SL) and Dempster-Shafer theory (DST) to parameterize the Dirichlet concentration distribution, and we develop a trainable uncertainty estimation calibration mechanism to optimize the boundary between certainty and uncertainty.

	\item[$\bullet$] We employ hybrid image encoder which better captures both the local and global information, generating multiple segmentation results to optimize predictions.

\end{itemize}

\section{Related Works}

\subsection{Interactive segmentation methods}
\begin{figure*}[t]
	\centering
	\includegraphics[width=\linewidth]{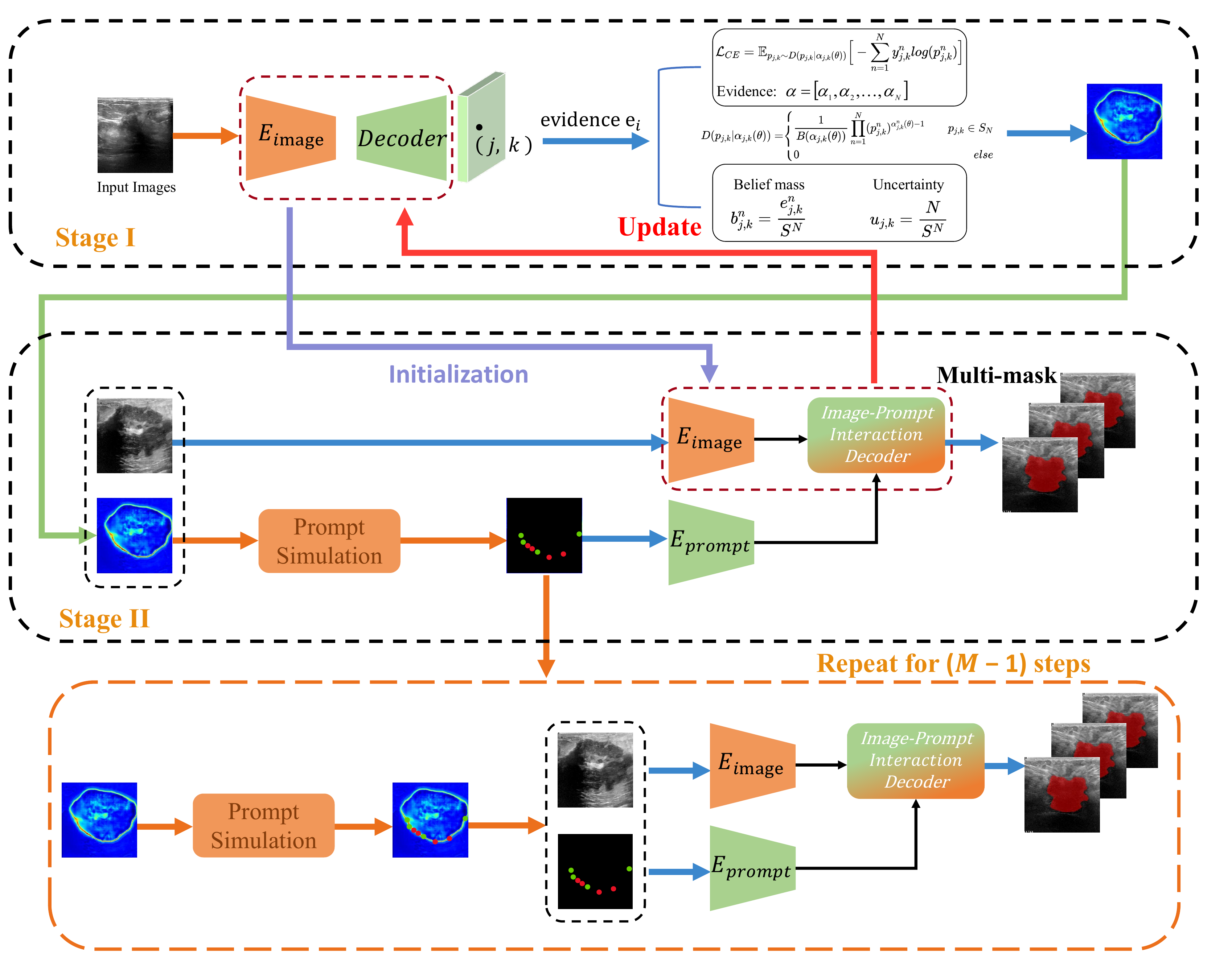}
	\caption{Overview of our two-stage interactive segmentation paradigm through evidence-based uncertainty estimation. In the first stage, an evidential calibrated model is trained to generate the uncertainty map, and it retains uncertainty map from the first-stage model are used as the point prompt generator, retain the interactive segmentation model in the second stage, which can be repeated for $M-1$ iterations in the process of forward propagation.}
	\label{fig:fangfa}
\end{figure*}
Interactive image segmentation is a longstanding research topic that has garnered significant attention, with an increasing number of works being proposed.
Traditional interactive segmentation methods\cite{graphcut, randomwalks, geos, grabcut} primarily focus on guiding segmentation at the pixel level through user interactions. These approaches leverage graph theory principles, representing image pixels as nodes in a graph and defining pixel similarity or connectivity through edges.
However, the aforementioned approaches solely concentrate on low-level image features, which exhibit limited generalization capabilities when dealing with complex objects and necessitate extensive user interaction.
With the further advancement of deep learning, semi-automatic\cite{banzidong1, banzidong3, banzidong2} interactive segmentation paradigm has demonstrated immense potential.
Semi-automatic methods enhance segmentation performance by harnessing the powerful capabilities of convolutional neural networks, complemented by limited user interaction.
Notably, the human-in-the-loop interaction process is akin to a double-edged sword; it provides targeted prompt information, yet the model may become excessively dependent on such specialized prompts,  increasing the complexity of training process. 
End-to-end\cite{masam, sammed2d, samu, desam, medsam, msa, samed} interactive segmentation methods aim to achieve end-to-end training process by automatically simulating the human prompted information.
Sampling regions where the model’s predictions are erroneous to generate prompt information can emulate human interactive behavior to a certain degree.
Nevertheless, the simulation process is not targeted and requires an extensive amount of guidance information, iterative processes, and training epochs to achieve improved performance.
In contrast, our model is designed to emulate the expert interactive behavior in a more natural manner and to integrate the benefits of both semi-automatic and end-to-end approaches, utilizing limited prompt information to realize end-to-end training and attain satisfactory performance.

% \vspace{-2.5mm}

\subsection{Uncertainty estimation}
Uncertainty estimation has been extensively applied to image segmentation tasks. 
These methods can be broadly categorized into three types: (1) Bayesian inference techniques\cite{gailv6, gailv4, gailv5, gailv1, gailv3, gailv2, uncercsvt, uncercsvt2, uncercsvt3}; (2) Deep ensemble-based approaches\cite{jicheng2, jicheng4, jicheng1, jicheng3}; (3) Evidence-based methods\cite{evi4, evi1, evidentialdeeplearining, evi3, evi2, evi5}.
% Specifically, bayesian inference techniques estimate the uncertainty from computing the posterior distribution over parameters or entropy on the training samples.
For instance, zhou et al.\cite{uncercsvt} leverages the Shannon-entropy to estimate the uncertainty for each location from images.
Monte Carlo dropout\cite{gailv4, gailv3, gailv2} leverages dropout during inference to approximate Bayesian posterior sampling, providing a practical measure of uncertainty, while conditional variational autoencoders\cite{gailv6, gailv5, gailv1} use a Bayesian paradigm to encode data variability into a latent space, offering a structured approach to uncertainty estimation.
Deep ensemble-based approaches\cite{jicheng2, jicheng4, jicheng1, jicheng3} derive uncertainty assessments from the collective outputs of multiple models, enhancing neural network robustness by aggregating insights from diversely trained models, with each contributing distinct perspectives to the overall uncertainty quantification.
However, the principal limitations of the aforementioned methods are rooted in their extensive computational costs, substantial memory requirements, and overall resource intensity.
Recently, evidence-based methods have been advanced for uncertainty estimation in deep learning, offering a computationally efficient alternative to Bayesian and ensemble approaches.
These methods\cite{evi4, evi1, evidentialdeeplearining, evi3, evi2, evi5} use the outputs of neural networks as evidence to parameterize Dirichlet distribution, which captures the uncertainty across multiple classes. This approach not only streamlines the computation but also enhances the interpretability of segmentation results, crucial for clinical decision-making.
Inspired by these methods, we propose a new paradigm that combines interactive segmentation with uncertainty estimation.

\section{Methodology}

\subsection{Overview}
The overview of our EUGIS paradigm is displayed in Fig.\ref{fig:fangfa}.
Within this paradigm, we establish two stages: the first of which is dedicated to generating high-quality uncertainty maps, while the second stage focuses on an efficient interactive segmentation method guided by these uncertainty maps.
According to the Fig. \ref{fig:fangfa}, in the stage I, we employ the hybrid encoder-decoder architecture $f_\theta$ to produce evidence, which is then utilized to integrate the Dempster-Shafer theory (DST)\cite{DST} and Subjective Logic (SL)\cite{sl} to estimate the uncertainty scores.
In order to better clear the boundary between certainty and uncertainty, we calibrate the evidence-based uncertainty estimation by assigning high uncertainty to regions where predictions are inaccurate, and demonstrating high confidence in the opposite scenario.
The omprehensive exposition in training for evidential calibrated learning will be offered in Sec. \ref{uncermap}.

Furthermore, the iterative training process for interactive segmentation in stage II utilizes the identical paradigm as that employed in stage I.
The only difference is that we add the prompt encoder to the hybrid encoder-decoder architecture $f_\theta$, which can be defined as $f_\theta^{prompt}$.
We utilize the high-quality uncertainty map generated from stage I to simulate and create more targeted prompts.
As the weights of $f_\theta^{prompt}$ are updated, so too is the uncertainty map, ensuring that the prompts evolve in a manner that reflects the interactive behaviors of expert clinicians with greater flexibility and relevance.
We will delve into the details of iterative training in Sec. \ref{iterationII}.

\subsection{Evidence-based uncertainty estimation in stage I}
\label{uncermap}
\subsubsection{Evidential uncertainty theory and modeling}
\label{dilikelei}
In order to model uncertainty quantifications in stage I, we incorporate the Dempster-Shafer theory (DST)\cite{DST} and Subjective Logic (SL)\cite{sl} into our paradigm. 
Although Bayesian theory allows for the quantification of uncertainty by updating beliefs based on prior knowledge and observed data, it relies on explicitly defined prior distributions.
The DST is regarded as a generalization of the Bayesian theory to subjective probabilities, enabling the allocation of belief mass across multiple hypotheses rather than merely assigning probabilities to a single hypothesis. 
By not adhering to the additivity principle of probability theory, DST allows the sum of belief masses to be less than one, thereby providing a clear representation of the state of ``I don't know''.
In other words, DST enhances the capability of Bayesian theory to handle uncertainty, enabling effective quantification even with insufficient evidence.
SL further enriches this paradigm by formalizing belief assignments in DST as a Dirichlet distribution\cite{dirichletuncertainty} within a discriminative paradigm, facilitating the quantification of belief mass and uncertainty.

Specifically, for a evidence-based uncertainty estimation segmentation task with $N$ classes in stage I, given the input $\mathbf{X}\in{\mathbb{R}^{N \times H \times W \times C}}$, where $H$, $W$ and $C$ refer to the height, width and channel,
we propose a segmentation model $f_{\theta}$ with hybrid image encoder and mask decoder to obtain evidence vector $e_i$ of sample $x_i$:
\begin{equation}
	\label{evidence}
	e_i = \mathcal{B}(f_{\theta}(x_i)),
\end{equation}
where $f_{\theta}(x_i)$ represents the output logits of segmentation model $f_{\theta}$ for sample $x_i$.
$\mathcal{B}(\cdot)$ is an activation function that ensures the output of segmentation model $f_{\theta}$ is non-negative, thereby transforming the output logits $f_{\theta}(x_i)$ into evidence $e_i$.
In the realm of activation functions $\mathcal{B}(\cdot)$, a variety of options present themselves, including $ReLU(\cdot)$, $SoftPlus(\cdot)$ and $exp(\cdot)$.
For our paradigm, we have elected to employ $ReLU(\cdot)$ as the non-negative activation function.

Furthermore, the SL provides a belief mass $\textbf{b}_i = \sum_{n=1}^N b_i^n$ and an uncertainty mass $\textbf{u}_i$ for different classes of segmentation result for sample $x_i$.
% The $N + 1$ mass values are all non-negative and their sum is 1.
More specifically, for any given $(j, k)\text{-}th$ pixel from sample $x_i$, The $N + 1$ mass values are all non-negative and satisfy the following definition:
\begin{equation}
	\label{sum1}
	u_{j, k} + \sum_{n=1}^{N}b_{j, k}^{n} = 1,
\end{equation}
where $u_{j, k} \in \textbf{u}_i \geq 0$  and $b_{j, k}^{n} \in \textbf{b}_i \geq 0$, they represent the belief strength of the any given $(j, k)\text{-}th$ pixel in the $n\text{-}th$ class and the overall uncertainty score for the $(j, k)\text{-}th$ pixel in sample $x_i$, respectively.
Then, $b_{j, k}^{n}$ can be computed by leveraging the generated evidence $e_{j, k}^{n}$ from Eq. \ref{evidence}, the belief mass $b_{j, k}^{n}$ and uncertainty score $u_{j, k}$ of the $(j, k)\text{-}th$ pixel can be denoted as:
\begin{equation}
	\label{bu}
	b_{j, k}^{n} = \frac{e_{j, k}^{n}}{S^N},\quad  u_{j, k} = \frac{N}{S^N},
\end{equation}
where $S^N = \sum_{n=1}^{N}(e_{j, k}^{n} + 1)$ refers to the Dirichlet strength.
It can be seen from Eq. \ref{sum1} and Eq. \ref{bu} that the computed uncertainty score is always in contradiction with total evidence.
When no evidence is present, namely $\sum_{n=1}^{N}e_{j, k}^{n}=0$, the belief mass $b_{j, k}^{n}$ assigns to any given $(j, k)\text{-}th$ pixel across each class within sample $x_i$ is zero, resulting in an uncertainty score of 1.
In contrast, an infinite amount of evidence will eliminates the possibility of uncertainty, which means that the model has high confidence for the predictions.

In the paradigm of SL, the belief mass assignment of DST is formalized as a Dirichlet distribution with parameters $\boldsymbol{\alpha}_{j,k} = [\alpha_{j,k}^1, \alpha_{j,k}^2, \dots, \alpha_{j,k}^N]$, where $\alpha_{j,k}^n = e_{j,k}^n + 1$.
That is to say, the belief mass $b_{j, k}^{n}$ and uncertainty score $u_{j, k}$ can be computed by leveraging $b_{j, k}^{n} = (\alpha_{j,k}^n-1) / \sum_{n=1}^{N}\alpha_{j,k}^n $ and $u_{j, k}=N/\sum_{n=1}^{N}\alpha_{j,k}^n$ when the corresponding Dirichlet distribution with determined $\boldsymbol{\alpha}_{j,k}$ is provided.
Then, the Dirichlet distribution density function can be defined as:

\begin{equation}
	\begin{aligned}
		\label{Dirichlet}
		D(p_{j,k}|\alpha_{j,k}(\theta))=&\left\{
		\begin{aligned}
			& \frac{1}{B(\alpha_{j,k}(\theta))}\prod_{n=1}^{N}(p_{j,k}^n)^{\alpha_{j,k}^n(\theta)-1}   & p_{j,k} \in S_N,\\
			& 0   & else ,\\
		\end{aligned}
		\right. \\
		\mathcal{B}(\alpha_{j,k}(\theta)) =& \frac{\prod^{N}_{n=1}\Gamma(\alpha_{j,k}^{n}(\theta))}{\Gamma(\sum_{n=1}^{N}\alpha_{j,k}^{n}(\theta))},
	\end{aligned}
\end{equation}
where $\textbf{p}_{j,k}$ is a probability vector of length $k$, namely $\textbf{p}_{j,k} = (p_{j,k}^1, p_{j,k}^2, \dots, p_{j,k}^N)$, and $S_N = \{\textbf{p}_{j,k} | \sum_{n=1}^{N}  p_{j,k} =1 \ and \ p_{j,k} \geq 0  \}$ denotes the $N\text{-}$dimension simplex.
$B(\alpha_{j,k}(\theta))$ refers to $N\text{-}$dimension multinomial Beta function, which is the normalization constant of the Dirichlet distribution, ensuring that the integral of the probability density function over the entire probability simplex is equal to 1.

\subsubsection{Calibrated evidential uncertainty}
Although the paradigm of Dempster-Shafer theory (DST) and Subjective Logic (SL) offers an approach for directly learning evidential uncertainty, it may face issues arising from insufficient uncertainty calibration, which can obscure the boundary between certainty and uncertainty. 
For instance, a model that assigns excessive confidence to an inaccurate prediction or insufficient confidence to an accurate one can negatively impact the reliability of the model.
Consequently, a reliable and well-calibrated evidential uncertainty estimation approach should exhibit high uncertainty when the prediction of the model is inaccurate, and demonstrate confidence in the opposite scenario.
To achieve this, we introduce a novel loss function, termed CEU, designed to calibrate evidential uncertainty, thereby further clarifying the boundaries between certainty and uncertainty.
The loss function CEU can be defined as follows:
\begin{equation}
	\begin{aligned}
		\label{CEU}
		\mathcal{L}_{CEU} = - \Bigg\{\alpha_t \bigg[ \sum_{j,k \in \{ \hat{y}_{j,k} = y_{j,k} \}}  \sum_{n=1}^{N} b^n_{j,k}  log(1-u_{j,k}) \bigg]  \Bigg\}  \\
		- \Bigg\{(1-\alpha_t) \bigg[ \sum_{j,k \in \{ \hat{y}_{j,k} \neq y_{j,k} \}}  \sum_{n=1}^{N} (1-b^n_{j,k})  log(u_{j,k}) \bigg]  \Bigg\} .
	\end{aligned}
\end{equation}
In the initial stages of training a neural network, the optimization of the model is directed by exponential decay strategy.
$\alpha_t$ refers to the annealing factor, which can be defined by $\alpha_t = \alpha_0e^{-\frac{t}{T}}$, $t$ and $T$ represent the current epoch and the total epochs, respectively.
Therefore, CEU assigns higher penalties to inaccurate but confident pixels during the early stages of training, and to accurate but uncertain pixels during the later stages of training. This approach enables the model to exhibit higher uncertainty in the presence of inaccurate predictions and lower uncertainty when making accurate predictions.

\subsubsection{Evidential calibrated model Learning}
Dirichlet-based evidential models interpret the categorical output of a pixel from a sample as a probability distribution, allowing for several possible predictions to be generated, each associated with distinct probabilities.
To align the predicted probabilities $p_{j,k}$ more closely with the actual ground truth $y_{j,k}$, we employ the cross-entropy-based loss.
Since $\textbf{p}_{j,k}$ is sampled from a Dirichlet distribution, which in turn is determined by parameter $\boldsymbol{\alpha}_{j,k}(\theta)$, and $\boldsymbol{\alpha}_{j,k}(\theta)$ is a function of the model parameters $\theta$, it is infeasible to directly optimize this cross-entropy. 
This is due to the non-differentiable nature of the sampling operation, as well as the inherent randomness in the sampled $\textbf{p}_{j,k}$, which complicates the accurate assessment of the obtained $\boldsymbol{\alpha}_{j,k}(\theta)$.

Therefore, we adopt the concept of Type II Maximum Likelihood Estimation, where although we cannot directly optimize the cross-entropy-based loss, we can optimize the expectation of the cross-entropy-based loss with respect to the Dirichlet distribution, it can be defined as follows:
\begin{equation}
	\begin{aligned}
		\label{celoss}
		\mathcal{L}_{CE} &= \mathbb{E}_{p_{j,k} \sim D(p_{j,k}|\alpha_{j,k}(\theta))}\Big[-\sum_{n=1}^{N}y_{j,k}^{n}log(p_{j,k}^n)\Big]  ,
		% & = \int [\sum_{n=1}^{N}-y_{j,k}^{n}log(p_{j,k}^n)] \cdot D(p_{j,k}|\alpha_{j,k})  dp_{j,k} \\
		% & = \int \Big[\sum_{n=1}^{N}-y_{j,k}^{n}log(p_{j,k}^n)\Big] \cdot \frac{1}{B(\alpha_{j,k})}\prod_{n=1}^{N}p_{j,k}^{\alpha_{j,k}-1} dp_{j,k} \\
		% & = 
	\end{aligned}
\end{equation}
as mentioned in Sec. \ref{dilikelei}, our Dirichlet distribution is defined on the probability simplex, hence each component $p_{j,k}^n$ of $\textbf{p}_{j,k}$ follows a beta distribution, that is:
\begin{equation}
	\begin{aligned}
		\label{beta}
		p_{j,k}^n \sim Beta(\alpha_{j,k}^n, S^N - \alpha_{j,k}^n),
	\end{aligned}
\end{equation}
where $S^N = \sum_{n=1}^{N}\alpha_{j,k}^n$ refers to the Dirichlet strength.
The probability density function is:
\begin{equation}
	\begin{aligned}
		\label{betafunc}
		&\rho(p_{j,k}^{n}|\theta) = \frac{1}{\mathcal{B}(\alpha_{j,k}^n, S^N - \alpha_{j,k}^n) }(p_{j,k}^{n})^ {\alpha_{j,k}^n-1}(1-p_{j,k}^{n})^{S^N - \alpha_{j,k}^n-1} ,\\
		&\mathcal{B}(\alpha_{j,k}^n, S^N - \alpha_{j,k}^n) = \frac{\Gamma(\alpha_{j,k}^n)\cdot\Gamma(S^N - \alpha_{j,k}^n))}{\Gamma(\alpha_{j,k}^n + S^N - \alpha_{j,k}^n))}\\
		&\quad \quad \quad \quad \quad \quad \quad  =  \frac{\Gamma(\alpha_{j,k}^n)\cdot\Gamma(S^N - \alpha_{j,k}^n))}{\Gamma(S^N )},
	\end{aligned}
\end{equation}
where $\mathcal{B}(\cdot, \cdot)$ and $\Gamma(\cdot)$ are Beta and Gamma function, respectively.

Combining Eq. \ref{celoss}, Eq. \ref{beta} and Eq. \ref{betafunc}, $\mathcal{L}_{CE}$ can be computed as:
\begin{equation}
	\begin{aligned}
		\label{celossfinal}
		\mathcal{L}_{CE} &= -\sum_{n=1}^{N}y_{j,k}^{n} \int log(p_{j,k}^{n}) \cdot \rho(p_{j,k}^{n}|\theta) dp_{j,k}^{n} \\
		&= -\sum_{n=1}^{N}y_{j,k}^{n} \cdot \mathbb{E}_{p_{j,k}^n \sim Beta(\alpha_{j,k}^n, S^N - \alpha_{j,k}^n)}[log(p_{j,k}^{n})]\\
		&=\sum_{n=1}^{N}y_{j,k}^{n} \cdot [\psi(S^N)-\psi(\alpha_{j,k}^n)],
	\end{aligned}
\end{equation}
where $\psi$ refers to the digamma function, $y^n_{j,k}$ is the true label for class $n$ in any given $(j, k)\text{-}th$ pixel from the sample $x_i$. 
Analogously, the expectation of $p_{j,k}^{n}$, namely posterior probability, can be obtained by integrating over the marginal distribution of the Dirichlet distribution,  so that the Bayes risk of soft dice loss can be acquired.
This process of the expectation of $p_{j,k}^{n}$ can be calculated as:
% we can compute the expectation of $p_{j,k}^{n}$ as the predicted probability by leveraging the given evidence $e_{j,k}^n$, so that the Bayes risk of dice loss can be obtained.
\begin{equation}
	\begin{aligned}
		\label{dice1}
		\hat{p}_{j,k}^{n} =& \int p(y=n|p_{j,k}) \cdot D(p_{j,k}|\alpha_{j,k}(\theta))dp_{j,k} \\
		=& \int p_{j,k}^n \cdot \rho(p_{j,k}^{n}|\theta)dp_{j,k}^n\\
		=& \int p_{j,k}^n \cdot \frac{(p_{j,k}^{n})^ {\alpha_{j,k}^n-1}(1-p_{j,k}^{n})^{S^N - \alpha_{j,k}^n-1}}{\mathcal{B}(\alpha_{j,k}^n, S^N - \alpha_{j,k}^n) }dp_{j,k}^n\\
		=& \frac{1}{\mathcal{B}(\alpha_{j,k}^n, S^N - \alpha_{j,k}^n) }\int (p_{j,k}^{n})^ {\alpha_{j,k}^n}(1-p_{j,k}^{n})^{S^N - \alpha_{j,k}^n-1}dp_{j,k}^n\\
		=& \frac{\Gamma(S^N )}{\Gamma(\alpha_{j,k}^n)\cdot\Gamma(S^N - \alpha_{j,k}^n))} \cdot 
		\frac{\Gamma(\alpha_{j,k}^n+1)\cdot\Gamma(S^N - \alpha_{j,k}^n))}{\Gamma(S^N +1)}\\
		=& \frac{\Gamma(S^N ) \cdot \Gamma(\alpha_{j,k}^n+1)  }{\Gamma(\alpha_{j,k}^n)\cdot \Gamma(S^N +1)}\\
		=& \frac{\alpha_{j,k}^n}{S^N},
	\end{aligned}
\end{equation}
where $\hat{p}_{j,k}^{n}$ represents the posterior probability in the $n\text{-}th$ class corresponding to the Dirichlet distribution.
Then the Bayes risk of soft dice loss can be defined as follows:
\begin{equation}
	\begin{aligned}
		\label{dice2}
		\mathcal{L}_{Dice} = 1- \sum_{j,k \in x_i} \frac{2 \cdot y_{j,k}^{n} \cdot \hat{p}_{j,k}^{n} + \beta_1}{y_{j,k}^{n} + \hat{p}_{j,k}^{n} + \beta_2}.
	\end{aligned}
\end{equation}
In addition, to ensure that the model can effectively express the uncertainty of predictions when sufficient evidence is lacking and to prevent evidence collapse, we incorporate Kullback–Leibler (KL)-divergence that can be defined as:
\begin{equation}
	\begin{aligned}
		\label{kl}
		\mathcal{L}_{KL} = &KL\Big[D(\textbf{p}_{j,k}|\widetilde{\boldsymbol{\alpha}}_{j,k})||D(\textbf{p}_{j,k}|\textbf{1})\Big]\\
		=& \mathbb{E}_{p_{j,k} \sim D(p_{j,k}|\widetilde{\alpha}_{j,k})}\Bigg[log\frac{D(\textbf{p}_{j,k}|\widetilde{\boldsymbol{\alpha}}_{j,k})}{D(\textbf{p}_{j,k}|\textbf{1})}\Bigg]\\
		=& \mathbb{E}\Big[D(\textbf{p}_{j,k}|\widetilde{\boldsymbol{\alpha}}_{j,k})\Big] - \mathbb{E}\Big[D(\textbf{p}_{j,k}|\textbf{1})\Big] \\
		=& \mathbb{E}\Bigg[ -log\mathcal{B}(\widetilde{\boldsymbol{\alpha}}_{j,k}) + \sum_{n=1}^{N}(\widetilde{\alpha}_{j,k}^{n}-1) log p_{j,k}^n      \Bigg]\\
		-& \mathbb{E}\Bigg[ -log\mathcal{B}(\textbf{1}) + \sum_{n=1}^{N}(1-1) log p_{j,k}^n      \Bigg]\\
		=& -log\frac{\mathcal{B}(\textbf{1})}{\mathcal{B}(\widetilde{\boldsymbol{\alpha}}_{j,k})} + \mathbb{E}\Bigg[ 
		\sum_{n=1}^{N}  (\widetilde{\alpha}_{j,k}^{n}-1) log p_{j,k}^n           \Bigg]\\
		=& -log\frac{\mathcal{B}(\textbf{1})}{\mathcal{B}(\widetilde{\boldsymbol{\alpha}}_{j,k})} + \sum_{n=1}^{N}  (\widetilde{\alpha}_{j,k}^{n}-1) \Bigg[ \psi(\widetilde{\alpha}_{j,k}^{n})-\psi(\sum_{n=1}^{N}\widetilde{\alpha}_{j,k}^{n})\Bigg]\\
		= & log\Bigg[ \frac{\Gamma(\sum_{n=1}^{N}\widetilde{\alpha}_{j,k}^{n})}{\Gamma(N)\sum_{n=1}^{N}\Gamma(\widetilde{\alpha}_{j,k}^{n})}   \Bigg]\\
		+& \sum_{n=1}^{N}  (\widetilde{\alpha}_{j,k}^{n}-1) \Bigg[ \psi(\widetilde{\alpha}_{j,k}^{n})-\psi(\sum_{n=1}^{N}\widetilde{\alpha}_{j,k}^{n})\Bigg],
	\end{aligned}
\end{equation}
where $\widetilde{\boldsymbol{\alpha}}_{j,k} = y_{j,k} + (1-  y_{j,k}) \; \odot \boldsymbol{\alpha}_{j,k}$ refers to the adjusted parameters of the Dirichlet distribution, $D(\textbf{p}_{j,k}|\textbf{1})$ is the Dirichlet distribution with uniform parameters,
which can encourage model to diminish the evidence for incorrect classes.

Finally, the overall loss function of evidential calibrated model learning can be defined as follows:
\begin{equation}
	\begin{aligned}
		\label{totalloss}
		\mathcal{L}_{Ecml}= \mathcal{L}_{CE} + \mathcal{L}_{CEU} + \lambda_1\mathcal{L}_{KL} + \lambda_2\mathcal{L}_{Dice},
	\end{aligned}
\end{equation}
where $(\lambda_1, \lambda_2)$ are the balance factors.

\subsection{Iterative training in stage II for EUGIS}
\label{iterationII}
\subsubsection{Problem formulation}
\label{multimask}
Consider $t$ as an interactive segmentation task that consisting of pairs of image and true label, $\{(x^t, y^t)_{i}\}_{i=1}^{N}$.
At the iteration $m$ in one epoch, given an ultrasound image $x^t$ and a set of user interactive behaviors $c_m$, we need to learn function $f_{\theta}^{prompt}(x^t, c_m)$ with parameters $\theta$, which can produce $K$ segmentation results $[\hat{y}_m^1, \hat{y}_m^2, \dots, \hat{y}_m^K]$.
The set of user interactive behaviors $c_m$ refers to the positive or negative points (clicks).

We first compute the difference between the true segmentation label $y^t$ and each of the segmentation results $\hat{y}_m^1, \hat{y}_m^2, \dots, \hat{y}_m^K$, so that we can choose the segmentation result with highest confidence score to optimize predictions and enhance robustness. 
It can be defined as:
\begin{equation}
	\label{multiseg}
	\mathcal{A}_{high\textendash con} = \mathbb{E}_{ \{ \hat{y}_m^k \in \hat{y}_m^K   \}} \Big[\mathcal{L}_{con}(y^t, \hat{y}_m^k) \Big],
\end{equation}
where $\mathcal{L}_{con}$ represents the mean square error loss, $\mathcal{A}_{high\textendash con}$ refers to the highest confidence score among these $K$ segmentation results in comparison to the true segmentation label $y^t$.

Then we minimize the difference between the true segmentation label $y^t$ and every iterative predictions with highest confidence score $[\hat{y}_1, \dots, \hat{y}_M]$, this process can be represented as: 
\begin{equation}
	\begin{aligned}
		\label{interseg}
		\mathcal{L}_{\theta} = \mathbb{E}_{\{(x^t, y^t) \in  t \}} \Bigg[  \mathbb{E}_{ \{ 
			\mathcal{L}_{con}(y^t, f_{\theta}^{prompt}(x^t, c_m)) = \mathcal{A}_{high\textendash con} 
			\}}  \Bigg[
		\\    \sum_{m=1}^{M} \mathcal{L}_{seg}(y^t, f_{\theta}^{prompt}(x^t, c_m)) \Bigg]  \Bigg],
	\end{aligned}
\end{equation}
where $\mathcal{L}_{seg}$ is a supervised segmentation loss.

In the training phase, we simulate a set of click or point interactions $c_m$ based on uncertainty map in Sec. \ref{uncermap}, so that predict segmentation result $\hat{y}_m$, which can be repeated for $M$ iterations.
In the following sections, we will explain the strategies for simulating $c_m$ and how to implement the iterative learning to optimize $ \mathcal{L}_{\theta}$.

\subsubsection{Prompt simulation}
\label{simulate}
% 尽量给出一个示意图，给出uncertainty map然后在高不确定性区域点个点
At present,  the prompt simulation methods employed in end-to-end interactive segmentation generate prompts within the boundaries of predicted error regions in a stochastic manner.
These approaches lack specificity and flexibility, resulting in inefficiency and often requiring a greater number of prompts and iterations to achieve satisfactory performance. 
To implement a prompt simulation mechanism that can effectively simulate the interactive behaviors of expert clinicians, we propose utilizing uncertainty maps to guide the prompt simulation process.
The essence of this method lies in harnessing the high-quality uncertainty map outputted in stage I in Sec. \ref{uncermap}, which provides an intuitive representation of the uncertainty of model regarding its predictions.

In particular, given a sample $x_i$, we can obtain the uncertainty score $u_{j,k}$ for any given $(j, k)\text{-}th$ pixel from the uncertainty map in stage I.
Subsequently, during iteration $m$, we identify the $k$ pixels exhibiting the greatest uncertainty scores and sample these as point prompts, which are subsequently fed into the prompt encoder. 
This process is repeatable, allowing for $M$ iterations to refine the interactive segmentation result.

% with the highest uncertainty scores as point prompts to be input into the prompt encoder, a process that can be repeated $M$ times.

\subsubsection{Iterative learning approach}
In our EUGIS, the iterative learning approach in stage II is shown in Fig. \ref{fig:fangfa}.
The training phase in second stage is iterative and aims to refine the segmentation results progressively. 
This phase incorporates the uncertainty map generated in stage II to simulate the behavior of radiologists and generate point prompts, as described below:
\paragraph{\textbf{Initialization}}
At the beginning of stage II, the model $f_{\theta}^{prompt}$ adopts the image encoder and decoder weights from the model $f_{\theta}$ in stage I. The weights for the prompt decoder in $f_{\theta}^{prompt}$ are initialized randomly. 

\paragraph{\textbf{Iterative Training (Fig. \ref{fig:fangfa})}}
During the first iteration or step ($m = 1$), the iterative training process incorporates two primary inputs: (1) images; and (2) point prompts. These prompts are generated from the uncertainty map described in Stage I, as explained in Sec. \ref{simulate}. 
These inputs will be encoded and interacted through the image encoder and prompt encoder, and then $K$ segmentation results $[\hat{y}_m^1, \hat{y}_m^2, \dots, \hat{y}_m^K]$ will be generated by the decoder (Sec. \ref{multimask}). We select the segmentation result with the highest confidence score as the final segmentation mask for computing the loss function $\mathcal{L}_{\theta}$. For the remaining iterations, the image features are generated only once at the initial iteration.
The loss function in the stage II can be defined as:
\begin{equation}
	\begin{aligned}
		\mathcal{L}_{\theta} = \frac{(M \times \mathcal{L}_{seg}) + (\mathcal{L}_{CE} + \mathcal{L}_{CEU} + \lambda_1\mathcal{L}_{KL})}{M},
	\end{aligned}
\end{equation}
where $M$ denotes the number of iteration, the $\mathcal{L}_{seg}$ represents the dice loss between the ground truth $y^t$ and the output segmentation result from the model $f_{\theta}^{prompt}$. The $\mathcal{L}_{CE}$, $\mathcal{L}_{CEU}$ and $\mathcal{L}_{KL}$ are the evidential calibration loss. 

\paragraph{\textbf{Epoch Training (Fig. \ref{fig:fangfa})}}
After the completion of the first epoch, the updated parameters of the model $f_{\theta}^{prompt}$ are systematically reintroduced into the model $f_\theta$ used in stage I (only the image encoder and decoder). Model $f_\theta$ then executes forward propagation with the newly updated parameters, updating the uncertainty map. This new uncertainty map will serve as the input for the next epoch, generating new point prompts and ensuring that each epoch is optimized based on feedback from the previous one.

\section{Experiments}
\subsection{Datasets}
In order to assess the effectiveness of the EUGIS system, we undertook an exhaustive evaluation across three widely recognized public ultrasound datasets: BUSI\cite{busi}, DDTI\cite{ddti}, and EchoNet\cite{echo}. This evaluation spanned three distinct diagnostic tasks, each focusing on a different disease—namely, breast lesions, thyroid nodules, and cardiovascular.

% across three tasks involving different diseases, namely breast lesions, thyroid nodules, and left ventricle.

\textbf{BUSI:} The Breast Ultrasound Scan Images (BUSI) collection, procured in 2018, encompasses a total of 780 breast ultrasound images. These images were derived from a cohort of 600 women, whose ages ranged from 25 to 75 years, with an average image size of $500\times500$ pixels. The images are categorized into three classes: normal, benign, and malignant. Additionally, detailed segmentation annotations corresponding to the tumors are provided within the benign and malignant breast ultrasound images.
% The collection of 2018-procured Breast Ultrasound Scan Images (BUSI) encompasses a total of 780 breast ultrasound images, derived from a cohort of 600 women, ranging in age from 25 to 75 years, with an average image size of $500\times500$ pixels. The images are categorized into three classes: normal, benign, and malignant. Within the benign and malignant breast ultrasound images, there are also detailed segmentation annotations corresponding to the tumors.

\textbf{DDTI:} DDTI (Digital Database of Thyroid Images) is a publicly accessible database of thyroid ultrasound images, supported by the National University of Colombia, CIM@LAB, and IDIME (Instituto de Diagnostico Medico). It comprises 99 cases and 134 images, encompassing a variety of lesions such as thyroiditis, cystic nodules, adenomas, and thyroid cancer, along with accurate delineations of the lesions.

\textbf{EchoNet:} The EchoNet-Dynamic dataset encompasses a collection of 10,030 four-chamber apical echocardiography video recordings, which were derived from clinical examinations performed at Stanford Hospital from 2016 to 2018. These videos underwent preprocessing to remove unnecessary segments, were scaled down to dimensions of 112$\times$112 pixels, and were marked with the contours of the left ventricular endocardium at both end-systolic and end-diastolic phases. We employ  these annotated frames as references for segmentation to create a novel dataset aimed at the segmentation tasks of this research. This dataset for segmentation consists of 20,046 images, with each image having its own corresponding segmentation map.

\subsection{Implementation details}
Our EUGIS was implemented based on PyTorch \cite{paszke2019pytorch} and trained by leveraging a NVIDIA RTX 3090 server with a 24G GPU.
we applied some regular data augmentation techniques to the images, such as random rotation and flipping.
The enhanced image was adjusted to 256 $\times$ 256 pixels before being utilized as input to the model.
In the EchoNet dataset, we preserved the uniform resolution of 112 $\times$ 112 pixels, aligning with the native dimensions of the four-chamber apical echocardiography videos, which are also 112 $\times$ 112 pixels.
We evaluate the proposed method and other methods on all datasets using five-fold cross-validation.
simultaneously, we employed the AdamW as the optimizer with the original learning rate of 1e-4 for 100 epochs. 
During the initial and subsequent phases of training, we specified a fixed number of epochs: 50 for the first stage and 50 for the second.
Additionally, the factors $(\lambda_1, \lambda_2)$ of evidential calibrated model learning from training stage I were set to $(0.2, 1)$.
For all datasets, Dice Score (Dice), Jaccard (IOU) and 95$\%$ Hausdorff distance (95HD) were used as evaluation metrics.

\begin{table*}[ht]
	\begin{center}
		\caption{Quantitative comparison results on the BUSI dataset. The results in bold are the best. $^\dagger$ means non-interactive segmentation models. $^\ddag$ means interactive segmentation methods.}
		\label{tab:BUSI}
		\setlength{\tabcolsep}{8mm}{
			\begin{tabular}{ c | c | c  c  c  c }
				\toprule[1.3pt]
				Methods 
				% & Year 
				& Num. pt & Dice (\%) $\uparrow$ & Jaccard (\%) $\uparrow$  & 95HD (pixel) $\downarrow$ \\
				\midrule[0.6pt]
				
				U-Net ~\cite{unet}$^\dagger$ 
				% & 2015  
				& \textbf{--}   &  73.45 $\pm$ 4.24  &  64.23 $\pm$ 4.02  &  39.25 $\pm$ 6.88    \\
				% \hline
				U-Net++~\cite{unet++}$^\dagger$ 
				% & 2018
				& \textbf{--}  & 76.38 $\pm$ 3.17 & 67.43 $\pm$ 3.12 & 31.82  $\pm$  4.48    \\
				SegNet~\cite{segnet}$^\dagger$ 
				% & 2017
				& \textbf{--} &  74.33 $\pm$ 2.98  &  65.22 $\pm$ 2.66  & 35.54 $\pm$ 5.81   \\
				TransUNet~\cite{transunet}$^\dagger$ 
				% & 2021
				& \textbf{--} &67.08  $\pm$ 3.01 &55.51  $\pm$  3.04 &65.62 $\pm$  6.95  \\
				H2Former ~\cite{h2former}$^\dagger$ 
				% & 2023
				& \textbf{--} & 77.53 $\pm$ 3.59 & 69.28 $\pm$ 3.31 & 28.35  $\pm$  3.56  \\
				\midrule[0.6pt]
				SAMed~\cite{samed}$^\ddag$ 
				% & 2023
				& \textbf{--} & 81.87 $\pm$ 0.18 & 71.82 $\pm$ 0.24 & 9.66 $\pm$ 3.41   \\
				MedSAM~\cite{medsam}$^\ddag$
				% & 2024
				& \textbf{--} &78.73 $\pm$ 1.04 & 67.32 $\pm$ 1.27 & 46.97 $\pm$ 3.92   \\
				\midrule[0.6pt]
				\multirow{3}*{SAM~\cite{sam}$^\ddag$} 
				% & \multirow{3}*{2023} 
				& 1 click  & 64.13 $\pm$ 1.86 & 52.84 $\pm$ 1.49 & 29.25 $\pm$ 1.05   \\
				& 3 click &  66.79 $\pm$ 0.94  & 55.53 $\pm$ 1.56 & 28.91 $\pm$ 0.71   \\
				& 5 click  & 67.73 $\pm$ 0.86 & 56.36 $\pm$ 1.02 & 27.29 $\pm$  0.66  \\
				\midrule[0.6pt]
				\multirow{3}*{SAM\text{-}Med2D~\cite{sammed2d}$^\ddag$}
				% & \multirow{3}*{2023}
				& 1  click  &80.89  $\pm$ 0.45 & 71.27 $\pm$ 0.89 & 27.83 $\pm$  4.57 \\
				& 3  click  & 82.62 $\pm$ 0.53 & 73.19 $\pm$ 0.87 & 22.48 $\pm$ 3.15  \\
				& 5  click  & 83.73 $\pm$ 0.42 & 74.53 $\pm$ 0.72 & 15.74 $\pm$ 1.57  \\
				
				\midrule[0.6pt]
				\multirow{3}*{Medical SAM Adapter~\cite{msa}$^\ddag$}
				% & \multirow{3}*{2023}
				& 1 click  &76.19 $\pm$ 0.38  &64.41  $\pm$ 0.28 &18.26  $\pm$  3.04  \\
				& 3 click &78.39 $\pm$ 0.34 &67.51  $\pm$ 0.26 &16.48  $\pm$  1.73  \\
				& 5 click &78.81 $\pm$ 0.31 &67.85  $\pm$ 0.22 &15.42  $\pm$  0.38 \\
				\midrule[0.6pt]
				\multirow{3}*{EUGIS (Ours)} 
				% & \multirow{3}*{\textbf{--}} 
				& 1 click & \textbf{88.95 $\pm$ 0.19} & \textbf{ 81.08 $\pm$ 0.14 } & \textbf{4.28 $\pm$ 1.35}  \\
				& 3 click & \textbf{89.17 $\pm$ 0.16} & \textbf{81.34 $\pm$ 0.12} & \textbf{4.02 $\pm$ 1.27} \\
				& 5 click & \textbf{89.93 $\pm$ 0.12} & \textbf{83.07 $\pm$ 0.14} & \textbf{2.81 $\pm$ 1.21}  \\
				\bottomrule[1.3pt]
		\end{tabular}}
	\end{center}
\end{table*}

\begin{figure*}[t]
	\centering
	\includegraphics[width=\linewidth]{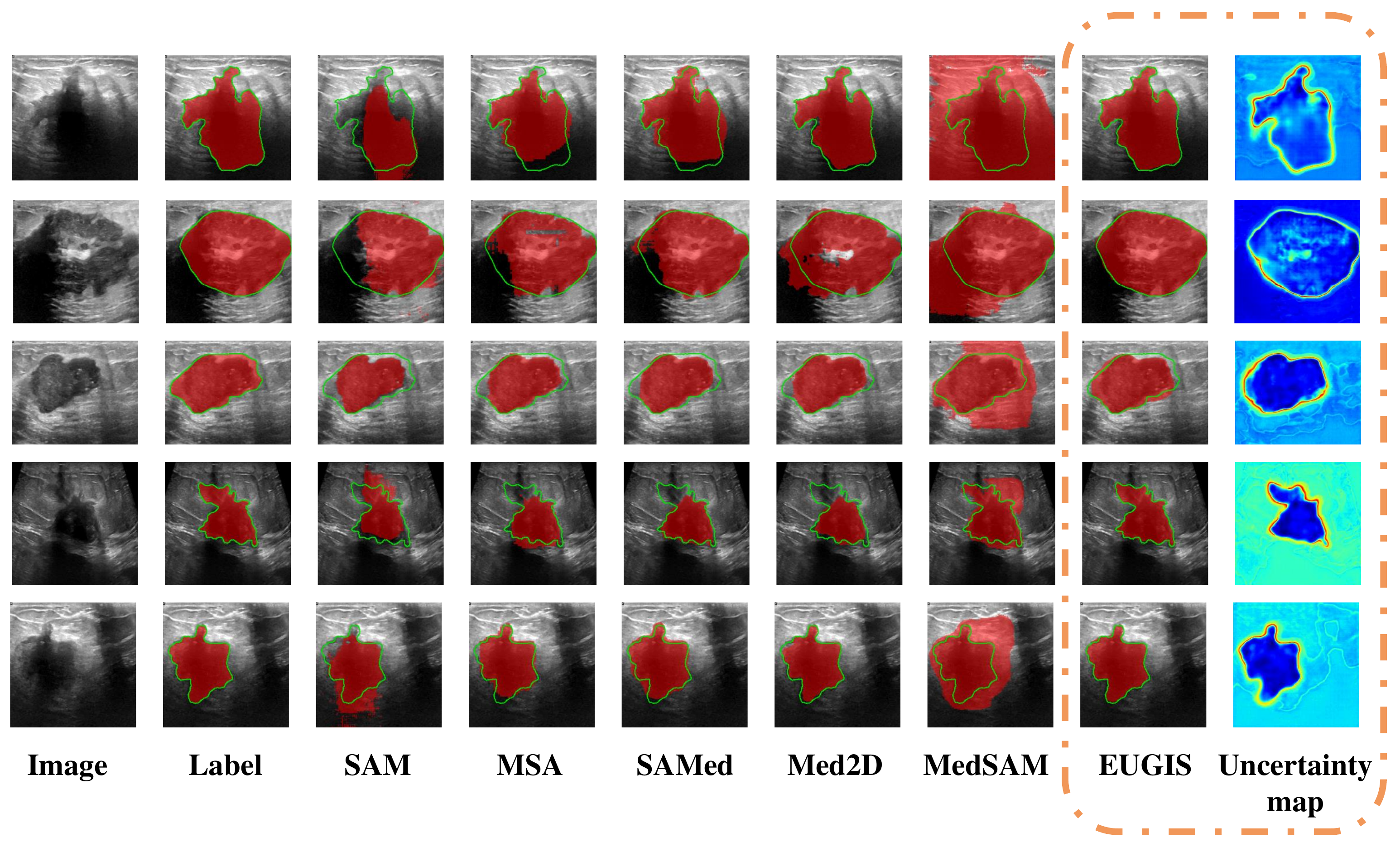}
	\caption{Visualization of comparison experimental results on the BUSI dataset. 
		The two columns on the far right represent our proposed method EUGIS and the corresponding uncertainty map.
		Label, MSA and Med2D refer to the Ground Truth, Medical SAM Adapter and SAM-Med2D, respectively. }
	\label{fig:keshihua}
\end{figure*}

\begin{table*}[ht]
	\begin{center}
		\caption{Quantitative comparison results on the DDTI dataset. The results in bold are the best. $^\dagger$ means non-interactive segmentation models. $^\ddag$ means interactive segmentation methods.}
		\label{tab:ddti}
		\setlength{\tabcolsep}{8mm}{
			\begin{tabular}{ c | c | c  c  c  c}
				\toprule[1.3pt]
				Methods 
				% & Year 
				& Num. pt & Dice (\%) $\uparrow$ & Jaccard (\%) $\uparrow$  & 95HD (pixel) $\downarrow$ \\
				\midrule[0.6pt]
				U-Net ~\cite{unet}$^\dagger$  
				% & 2015 
				& \textbf{--}   &  76.84 $\pm$ 4.09   & 65.14 $\pm$ 3.62  &  32.11 $\pm$ 3.83 \\
				% \hline
				U-Net++~\cite{unet++}$^\dagger$ 
				% & 2018
				& \textbf{--}  &  77.57 $\pm$ 3.57   &  64.46 $\pm$ 3.04  &  35.63 $\pm$ 4.99  \\
				SegNet~\cite{segnet}$^\dagger$
				% & 2017 
				& \textbf{--} &  75.31 $\pm$ 3.12    & 62.18 $\pm$ 2.71 &  36.39 $\pm$ 3.32  \\
				TransUNet~\cite{transunet}$^\dagger$ 
				% & 2021
				& \textbf{--} & 61.68 $\pm$ 3.62   & 46.79 $\pm$ 2.61  & 57.16 $\pm$ 3.28 \\
				H2Former ~\cite{h2former}$^\dagger$
				% & 2023 
				& \textbf{--} &  80.62 $\pm$ 2.91    &  68.53 $\pm$ 2.01  &  31.45 $\pm$ 2.04 \\
				\midrule[0.6pt]
				SAMed~\cite{samed}$^\ddag$ 
				% & 2023
				& \textbf{--} & 90.77 $\pm$ 0.62 & 83.17 $\pm$ 1.06  & 4.21 $\pm$ 0.56 \\
				MedSAM~\cite{medsam}$^\ddag$ 
				% & 2024
				& \textbf{--} & 73.05 $\pm$ 0.81& 58.76 $\pm$ 0.92  &  94.85$\pm$ 3.12 \\
				\midrule[0.6pt]
				\multirow{3}*{SAM~\cite{sam}$^\ddag$} 
				% & \multirow{3}*{2023} 
				& 1 click  & 50.69 $\pm$ 0.45   & 35.78 $\pm$ 0.51 & 61.35 $\pm$ 0.36 \\
				& 3 click & 62.89 $\pm$ 1.22&47.26 $\pm$ 1.20 & 52.81 $\pm$ 1.06   \\
				& 5 click  & 66.13 $\pm$ 0.36  &50.76 $\pm$  0.36&51.55  $\pm$ 0.67   \\
				\midrule[0.6pt]
				\multirow{3}*{SAM\text{-}Med2D~\cite{sammed2d}$^\ddag$}
				% & \multirow{3}*{2023}
				& 1  click  & 88.67 $\pm$  0.58  & 80.09 $\pm$ 1.11 & 27.67 $\pm$ 0.79 \\
				& 3  click  & 89.40 $\pm$ 0.34 & 81.38 $\pm$ 0.58 & 25.72 $\pm$ 1.03  \\
				& 5  click  & 89.97 $\pm$ 0.22 & 82.17 $\pm$ 0.36 & 24.93 $\pm$ 0.27   \\
				\midrule[0.6pt]
				% \multirow{3}*{MedSAM~\cite{medsam}$^\ddag$} & 1 click  &  $\pm$    &  $\pm$  & $\pm$  \\
				%  & 3 click  &  $\pm$  &  $\pm$  &  $\pm$    \\
				%  & 5 click  &  $\pm$  &  $\pm$  &  $\pm$    \\
				%  \midrule[0.6pt]
				\multirow{3}*{Medical SAM Adapter~\cite{msa}$^\ddag$}
				% & \multirow{3}*{2023}
				& 1 click  &89.71 $\pm$ 0.72   & 81.72 $\pm$ 1.09 & 5.49 $\pm$ 0.56 \\
				& 3 click &90.51 $\pm$ 0.91 & 82.84 $\pm$ 0.56 & 4.01 $\pm$  0.37  \\
				& 5 click &91.17 $\pm$ 0.69 &83.96  $\pm$  1.07&3.45  $\pm$ 0.28 \\
				\midrule[0.6pt]
				\multirow{3}*{EUGIS (Ours)} 
				% & \multirow{3}*{\textbf{--}} 
				& 1 click & \textbf{91.63 $\pm$ 0.31}  & \textbf{84.78 $\pm$ 0.52} & \textbf{3.67 $\pm$ 0.35} \\
				& 3 click & \textbf{91.75 $\pm$ 0.31} & \textbf{84.99 $\pm$ 0.51}  & \textbf{3.56 $\pm$ 0.34} \\
				& 5 click & \textbf{92.28 $\pm$ 0.29} & \textbf{85.86 $\pm$ 0.47}  & \textbf{3.01 $\pm$ 0.29} \\
				\bottomrule[1.3pt]
		\end{tabular}}
	\end{center}
\end{table*}

\begin{table*}[ht]
	\begin{center}
		\caption{Quantitative comparison results on the EchoNet dataset. The results in bold are the best. $^\dagger$ means non-interactive segmentation models. $^\ddag$ means interactive segmentation methods.}
		\label{tab:echo}
		\setlength{\tabcolsep}{8mm}{
			\begin{tabular}{ c | c | c  c  c  c }
				\toprule[1.3pt]
				Methods 
				% & Year
				& Num. pt & Dice (\%) $\uparrow$ & Jaccard (\%) $\uparrow$  & 95HD (pixel) $\downarrow$ \\
				\midrule[0.6pt]
				
				U-Net ~\cite{unet}$^\dagger$  
				% & 2015  
				& \textbf{--}   & 91.14 $\pm$ 1.34   & 84.18 $\pm$ 1.03 & 3.43 $\pm$ 0.68 \\
				% \hline
				U-Net++~\cite{unet++}$^\dagger$ 
				% & 2018 
				& \textbf{--}  &91.97  $\pm$ 0.74   & 85.51 $\pm$ 0.74 & 3.07 $\pm$ 0.52 \\
				SegNet~\cite{segnet}$^\dagger$  
				% & 2017
				& \textbf{--} & 89.16 $\pm$ 1.46   &83.23 $\pm$ 1.28 & 3.51 $\pm$ 0.57 \\
				TransUNet~\cite{transunet}$^\dagger$ 
				% & 2021 
				& \textbf{--} & 79.26 $\pm$ 1.06   & 70.69 $\pm$ 0.98 & 7.45 $\pm$ 0.53 \\
				H2Former ~\cite{h2former}$^\dagger$ 
				% & 2023 
				& \textbf{--} & 91.65 $\pm$ 0.96    & 84.98 $\pm$ 1.11 & 3.22 $\pm$ 0.51 \\
				\midrule[0.6pt]
				SAMed~\cite{samed}$^\ddag$
				% & 2023 
				& \textbf{--} & 91.06 $\pm$ 0.42   & 83.67 $\pm$ 0.25 & 2.43 $\pm$ 0.62 \\
				MedSAM~\cite{medsam}$^\ddag$ 
				% & 2024
				& \textbf{--} &71.99  $\pm$ 0.36&57.22  $\pm$ 0.48  & 11.13 $\pm$ 0.42 \\
				\midrule[0.6pt]
				\multirow{3}*{SAM~\cite{sam}$^\ddag$} 
				% & \multirow{3}*{2023}  
				& 1 click  & 71.47 $\pm$  2.18   & 58.57 $\pm$ 2.09 & 25.69 $\pm$ 1.97 \\
				& 3 click & 76.55 $\pm$ 1.99  & 63.39 $\pm$ 1.16 & 22.87 $\pm$ 1.73 \\
				& 5 click  & 78.28 $\pm$ 1.62   & 65.36 $\pm$ 0.86 & 21.73 $\pm$ 1.58 \\
				\midrule[0.6pt]
				\multirow{3}*{SAM\text{-}Med2D~\cite{sammed2d}$^\ddag$}
				% & \multirow{3}*{2023}
				& 1  click   & 90.99 $\pm$ 0.72 & 83.88 $\pm$ 0.67 & 4.48 $\pm$ 0.61 \\
				& 3  click  & 91.29 $\pm$ 0.63  & 84.41 $\pm$ 0.51 & 3.38 $\pm$ 0.42 \\
				& 5  click  & 91.83 $\pm$  0.48  & 85.26 $\pm$ 0.42 & 3.14 $\pm$ 0.37 \\
				\midrule[0.6pt]
				% \multirow{3}*{MedSAM~\cite{medsam}$^\ddag$} & 1 click  &  $\pm$    &  $\pm$  & $\pm$  \\
				%  & 3 click  &  $\pm$    &  $\pm$  & $\pm$  \\
				%  & 5 click  &  $\pm$    &  $\pm$  & $\pm$  \\
				% \midrule[0.6pt]
				\multirow{3}*{Medical SAM Adapter~\cite{msa}$^\ddag$}
				% &\multirow{3}*{2023}
				& 1 click  &90.83 $\pm$ 0.24   &83.41  $\pm$ 0.27 &1.63  $\pm$ 0.11 \\
				& 3 click &90.89 $\pm$ 0.19  &83.52  $\pm$ 0.21 &1.62  $\pm$ 0.08  \\
				& 5 click &91.11 $\pm$ 0.13  &83.84   $\pm$ 0.16 &1.41  $\pm$ 0.07 \\
				\midrule[0.6pt]
				\multirow{3}*{EUGIS (Ours)} 
				% &\multirow{3}*{\textbf{--}} 
				& 1 click & \textbf{94.85 $\pm$ 0.08}  & \textbf{90.35 $\pm$ 0.05} & \textbf{0.67 $\pm$ 0.04} \\
				& 3 click & \textbf{95.04 $\pm$ 0.04} & \textbf{90.68 $\pm$ 0.02}  & \textbf{0.62 $\pm$ 0.01} \\
				& 5 click & \textbf{95.13 $\pm$ 0.01} & \textbf{90.85 $\pm$ 0.02}  & \textbf{0.58 $\pm$ 0.01} \\
				\bottomrule[1.3pt]
		\end{tabular}}
	\end{center}
\end{table*}

\section{Results}

\begin{figure*}[t]
	\centering
	\includegraphics[width=\linewidth]{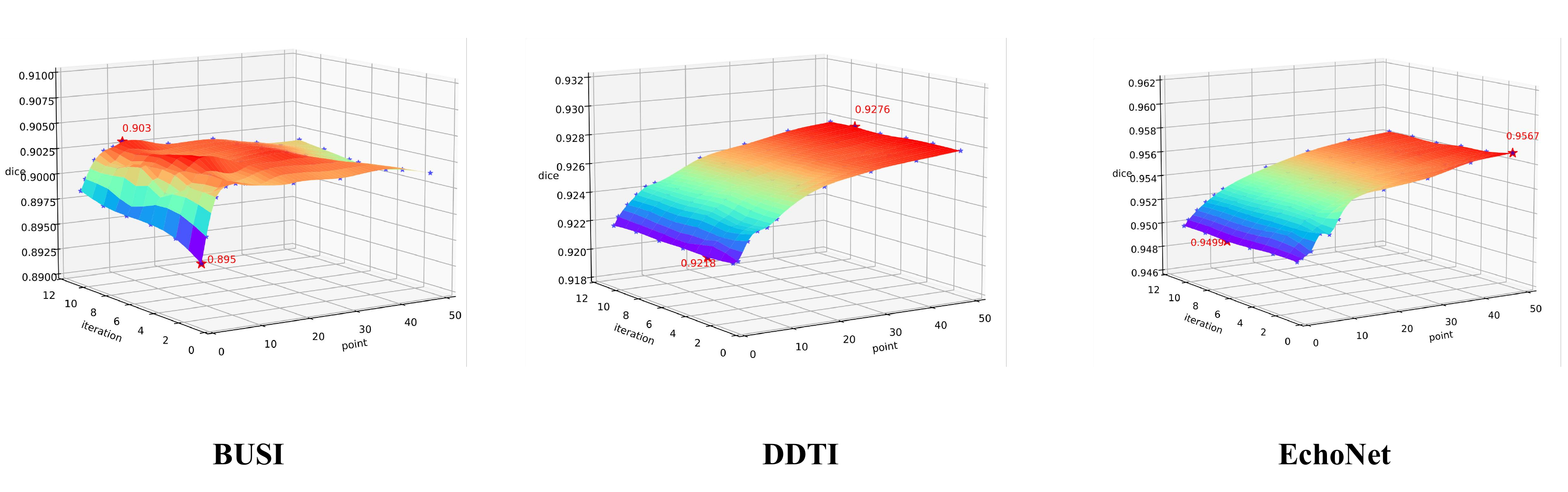}
	\caption{The effect of the number of point prompts and iterations on performance in EUGIS.}
	\label{fig:prompt}
\end{figure*}

\subsection{Comparison With Non-interactive and State-of-the-Art (SOTA)  Interactive Segmentation Methods}
In this section, we evaluate our proposed EUGIS against $5$ non-interactive segmentation learning approaches and advanced interactive segmentation models across $3$ distinct datasets.
These methods incorporate U-Net\cite{unet}, U-Net++\cite{unet++}, SegNet\cite{segnet}, TransUNet\cite{transunet}, H2Former\cite{h2former}, SAMed\cite{samed}, SAM\cite{sam}, SAM-Med2D\cite{sammed2d}, MedSAM\cite{medsam} and Medical SAM Adapter\cite{msa}.
To better facilitate comparison, we divide the interactive segmentation methods into two parts: the first consists of methods without point prompt, while the second comprises those with point prompt input\cite{sammed2d, sam,msa}.
Furthermore, for these interactive segmentation methods\cite{sammed2d, medsam,msa} that incorporate point prompt inputs,we also conducted a more in-depth comparison by utilizing varying quantities of point prompts ($1$, $3$ and $5$ point prompts).
Notably, these interactive segmentation models\cite{sammed2d,msa, samed} are all trained from scratch, rather than merely utilizing the pre-trained weights provided.

\subsubsection{Results on the BUSI dataset}
The quantitative results of all methods on the BUSI dataset are shown in Table \ref{tab:BUSI}. As expected, due to the relatively small size of the BUSI dataset, non-interactive segmentation methods failed to provide satisfactory results on the BUSI dataset. Among these, H2Former \cite{h2former} demonstrated the best performance, achieving average Dice and Jaccard scores of 77.53\% and 69.28\%, respectively. When compared to our newly proposed EUGIS, H2Former exhibited a significant performance gap of approximately 10 percentage points. Furthermore, non-interactive segmentation methods exhibited high standard deviations in their metrics, indicating a lack of generalization performance and robustness. Additionally, the performance of SAM \cite{sam} was less than satisfactory. Even with the assistance of point prompts, SAM struggled to handle the substantial domain shift between natural and medical images.

In contrast, interactive segmentation methods trained from scratch exhibited robustness with low standard deviations and performance that significantly surpassed that of non-interactive segmentation methods. Among the existing approaches, when both SAM-Med2D \cite{sammed2d} and Medical SAM Adapter \cite{msa} were given only one point prompt, SAMed \cite{samed} demonstrated commendable performance, second only to the proposed method EUGIS, with an average Dice of 81.87\% and Jaccard of 71.82\%. This suggests that the low-rank fine-tuning strategy of SAMed conferred an advantage over models with a limited number of point prompts.

However, as the number of point prompts increased, SAM-Med2D achieved the second-best performance. With five point prompts, the average Dice and Jaccard of SAM-Med2D reached 83.73\% and 74.53\%, respectively. Furthermore, our proposed EUGIS achieved remarkable performance with a single point prompt, attaining Dice and Jaccard scores of 87.69\% and 79.31\%, respectively, with standard deviations under 0.2\%. It demonstrated the best performance and generalization capability among all compared methods, outperforming the second-best SAM-Med2D \cite{sammed2d} (five-point prompts used) by nearly five percentage points.

In addition, as illustrated in figure \ref{fig:keshihua}, we visualized the segmentation results and uncertainty maps on the BUSI dataset, with the aim of gaining a clear and intuitive understanding of the performance differences between various methods and the distribution of confidence of segmentation result in our method. 
The red areas in the figure represented the ground truth or the segmentation results from different methods, while the green contours denoted the boundaries of the ground truth.
Notably, the visualization of SAM, SAM-Med2D, Medical SAM Adapter and EUGIS used only a single point prompt.
We intuitively observed  that the segmentation results of SAM and MedSAM were somewhat disorganized. 
SAM tended to under-segmentation, meaning it failed to fully capture the complete contours of the target, resulting in only a small portion of the target area being correctly segmented, which explained the high 95HD score of SAM.
In contrast, MedSAM leveraged bounding box (bbox) as the prompt, hence the resulting over-segmentation.
In other words, although MedSAM captured most of the target area, it also mistakenly segmented part of the background area, leading to confusion between the target and background regions, thereby achieving the highest 95HD score among all methods.
Moreover, by observing the uncertainty map generated by EUGIS, it was evident that the areas of segmentation error in EUGIS correspond to regions with high predicted uncertainty. 
For instance, in the segmentation results of EUGIS shown in the second row, it was clearly visible that it oversteps the segmentation boundary of the target area in the lower left corner, and this location corresponded precisely to the red area in the uncertainty map, as the red area indicated a higher level of predicted uncertainty.
Overall, EUGIS delivered the best visual segmentation results, successfully capturing most of the target regions while extracting their complete contours with relatively high precision.

\subsubsection{Results on the DDTI dataset}
We further conducted comparative experiments on another thyroid nodule ultrasound dataset DDTI, with the quantitative results presented in Table \ref{tab:ddti}.
We can easily observe that there is still a significant gap between non-interactive segmentation methods and interactive segmentation methods trained from scratch. This gap is reflected in both performance and generalization capabilities. 
Meanwhile, as SAM is not an interactive segmentation method trained from the ground up, its performance remains unsatisfactory.
However, among interactive segmentation methods trained from scratch, even though the dataset is smaller, these methods have demonstrated even stronger performance. 
Similar to the results on the BUSI dataset, with both SAM-Med2D and Medical SAM Adapter equipped with a single point prompt, SAMed still ranks second only to our proposed method, achieving an impressive average Dice and Jaccard of 90.77\% and 83.17\%, respectively.
Moreover, as the number of point prompts increases to five, Medical SAM Adapter, in turn, achieves performance second only to the proposed method, with an average Dice and Jaccard of 91.17\% and 83.96\%, respectively.
Despite the strong performance demonstrated by SAM-Med2D, Medical SAM Adapter, and SAMed, our proposed method, EUGIS, still outperforms all comparison methods even when using only a single point prompt. 
Thus, this is sufficient to demonstrate that our proposed method can achieve better segmentation results with fewer prompts.

\begin{table*}[ht]
	\begin{center}
		\caption{Ablation experiment results of the different encoder on the BUSI dataset. The results in bold are the best.}
		\label{tab:encoder}
		\setlength{\tabcolsep}{3.5mm}{
			\begin{tabular}{c c| c  c  c  }
				\toprule
				ViT Encoder & CNN Encoder  &  Dice (\%) $\uparrow$ & Jaccard (\%) $\uparrow$ & 95HD (pixel) $\downarrow$ \\
				\midrule
				$\checkmark$ &    & 84.28 $\pm$ 0.32 & 75.82 $\pm$ 0.29 & 6.31 $\pm$ 1.89\\
				& $\checkmark$   & 87.57 $\pm$ 0.21 & 79.43 $\pm$ 0.16 & 4.89 $\pm$ 1.47 \\
				$\checkmark$ &   $\checkmark$ &  \textbf{88.95 $\pm$ 0.19} & \textbf{ 81.08 $\pm$ 0.14 } & \textbf{4.28 $\pm$ 1.35}  \\
				\bottomrule
		\end{tabular}}
	\end{center}
\end{table*}

\begin{table*}[ht]
	\begin{center}
		\caption{Ablation experiment results of the effect in calibrated evidential uncertainty and multiple sgementation results on the DDTI dataset. The results in bold are the best.}
		\label{tab:cue}
		\setlength{\tabcolsep}{5mm}{
			\begin{tabular}{ c  c | c  c  c  }
				\toprule
				CEU & Multi Seg  &  Dice (\%) $\uparrow$ & Jaccard (\%) $\uparrow$ & 95HD (pixel) $\downarrow$ \\
				\midrule
				$\checkmark$ &  &   91.59 $\pm$ 0.33 & 84.74 $\pm$ 0.54 & 3.69 $\pm$ 0.36\\
				& $\checkmark$ &   90.17 $\pm$ 0.36 & 83.16 $\pm$ 0.59 & 3.96 $\pm$ 0.41 \\
				$\checkmark$ & $\checkmark$  & \textbf{91.63 $\pm$ 0.31}  & \textbf{84.78 $\pm$ 0.52} & \textbf{3.67 $\pm$ 0.35} \\
				\bottomrule
		\end{tabular}}
	\end{center}
\end{table*}

\subsubsection{Results on the EchoNet dataset}
In addition to these smaller datasets, we also conducted comparative experiments on echocardiography dataset containing over 20,000 images. The quantitative results are shown in Table \ref{tab:echo}.
Surprisingly, non-interactive segmentation methods have outperformed these interactive segmentation methods trained from scratch to some extent.
For instance, compared to the SAM-Med2D \cite{sammed2d} method with five point prompts, U-Net++\cite{unet++} achieved Dice and Jaccard scores that were 0.14\% and 0.25\% higher, respectively.
By comparison, even with just a single point prompt, our proposed model EUGIS outperforms the best U-Net++, attaining Dice and Jaccard scores that are higher by 2.88\% and 4.84\%, respectively.
This precisely demonstrates the philosophy of deep learning: as the amount of data increases, it allows the gap between different approaches to be bridged to some extent.
On the other hand, this also demonstrates that with limited point prompts, the interactive segmentation method guided by the error region is far inferior to our proposed method.

\subsection{Impact of Point prompts and Iterations}
To investigate the influence of the number of point prompts and iterations on the performance of EUGIS, we designed a set of comparative experiments. 
As shown in Figure \ref{fig:prompt}, the experimental results from left to right correspond to the BUSI, DDTI, and EchoNet datasets, respectively. 
The number of point prompts is set to range from 1 to 50, while the number of iterations was set between 1 and 12.
We found that the BUSI dataset reaches performance saturation earlier as the number of point prompts and iterations rises, which may be attributed to the inherently more ambiguous boundaries present in the BUSI dataset. This is further evidenced by the 95HD metric in Table \ref{tab:BUSI}, where the overall 95HD values for the BUSI dataset are consistently higher than those for the DDTI and EchoNet datasets.
In contrast, for the DDTI and EchoNet datasets, overall performance steadily improves with the addition of point prompts and iterations, indicating that the point prompt information we provide is crucial. However, an excessive number of point prompts and iterations also causes the performance gains of the DDTI and EchoNet datasets to slow down, approaching a saturation point.
It is noteworthy that the impact of the number of point prompts appears to be more significant.
The performance improvement driven by a larger number of point prompts is more noticeable,  whereas the performance enhancement brought about by the increase in iterations is minimal.  
Therefore, we conclude that point prompts play a dominant role in enhancing the performance of EUGIS, while iterations serve a more supportive role.

% \subsection{Visualization of Experimental Results}

\subsection{Ablation Study}

An ablation study of model configurations is presented in Table \ref{tab:encoder} and Table \ref{tab:cue}.
To begin, we conducted an extensive experimental analysis of the encoder used by EUGIS on the BUSI dataset.
Our findings revealed that the CNN-based encoder outperforms the Vision Transformer (ViT) when used in isolation.
Moreover, when relying solely on the Vision Transformer backbone, the performance of EUGIS declines even more noticeably.
This suggests that with fewer training epochs, the ViT encoder, which lacks inductive bias and local perceptual capabilities, fails to fully demonstrate its global feature extraction potential.
Consequently, the adoption of a hybrid encoder architecture enables us to leverage the strengths of both encoder types, ensuring optimal performance.
Additionally, we evaluated the effectiveness of Calibrated Evidential Uncertainty (CEU) and multiple segmentation heads (Multi Seg) on the DDTI dataset.
Our findings showed that Multi Seg yielded only a marginal improvement, with the dice score dropping by just 0.04\% when Multi Seg was omitted, indicating its limited impact.
In contrast, CEU had a more substantial effect: without CEU, the performance of EUGIS declined, with dice and Jaccard scores decreasing by 1.46\% and 1.62\%, respectively.
This demonstrates that calibrated uncertainty can significantly enhance the overall performance of EUGIS.

\section{Conclusions}
This paper introduces a novel interactive segmentation paradigm, driven by evidence-based uncertainty estimation, designed for ultrasound image segmentation tasks.
We begin by modeling uncertainty for different classes of segmentation results to generate uncertainty maps by leveraging Dempster-Shafer Theory (DST) and Subjective Logic (SL).
we have developed the Calibrated evidential uncertainty (CEU) to better generate calibrated uncertainty for providing more accurate uncertainty maps.
Moreover, by leveraging the confidence provided by the uncertainty map to generate point prompts, we not only make the prompts more precise but also achieve better and more robust segmentation results with fewer prompts.
Additionally, we utilized a hybrid encoder combined with multiple segmentation heads to effectively capture both local and global features, while also enhancing the robustness of the model.
The experimental results on three public ultrasound image datasets demonstrate that our proposed method EUGIS surpasses existing non-interactive segmentation and interactive segmentation methods, achieving state-of-the-art performance with just a single point prompt.
These results validate the effectiveness of this approach and its potential for clinical application, while also providing new insights for interactive segmentation methods.

\section{References}

\bibliographystyle{IEEEtran}
\bibliography{mybibliography}

\begin{thebibliography}{10}
\providecommand{\url}[1]{#1}
\csname url@samestyle\endcsname
\providecommand{\newblock}{\relax}
\providecommand{\bibinfo}[2]{#2}
\providecommand{\BIBentrySTDinterwordspacing}{\spaceskip=0pt\relax}
\providecommand{\BIBentryALTinterwordstretchfactor}{4}
\providecommand{\BIBentryALTinterwordspacing}{\spaceskip=\fontdimen2\font plus
\BIBentryALTinterwordstretchfactor\fontdimen3\font minus
  \fontdimen4\font\relax}
\providecommand{\BIBforeignlanguage}[2]{{%
\expandafter\ifx\csname l@#1\endcsname\relax
\typeout{** WARNING: IEEEtran.bst: No hyphenation pattern has been}%
\typeout{** loaded for the language `#1'. Using the pattern for}%
\typeout{** the default language instead.}%
\else
\language=\csname l@#1\endcsname
\fi
#2}}
\providecommand{\BIBdecl}{\relax}
\BIBdecl

\bibitem{unet}
O.~Ronneberger, P.~Fischer, and T.~Brox, ``U-net: Convolutional networks for
  biomedical image segmentation,'' in \emph{Medical Image Computing and
  Computer-Assisted Intervention--MICCAI 2015: 18th international conference,
  Munich, Germany, October 5-9, 2015, proceedings, part III 18}.\hskip 1em plus
  0.5em minus 0.4em\relax Springer, 2015, pp. 234--241.

\bibitem{unet++}
Z.~Zhou, M.~M. Rahman~Siddiquee, N.~Tajbakhsh, and J.~Liang, ``Unet++: A nested
  u-net architecture for medical image segmentation,'' in \emph{Deep Learning
  in Medical Image Analysis and Multimodal Learning for Clinical Decision
  Support: 4th International Workshop, DLMIA 2018, and 8th International
  Workshop, ML-CDS 2018, Held in Conjunction with MICCAI 2018, Granada, Spain,
  September 20, 2018, Proceedings 4}.\hskip 1em plus 0.5em minus 0.4em\relax
  Springer, 2018, pp. 3--11.

\bibitem{csvt2}
H.~Qi, H.~Zhou, J.~Dong, and X.~Dong, ``Small sample image segmentation by
  coupling convolutions and transformers,'' \emph{IEEE Transactions on Circuits
  and Systems for Video Technology}, 2023.

\bibitem{csvt4}
H.~Li, D.-H. Zhai, and Y.~Xia, ``Erdunet: An efficient residual double-coding
  unet for medical image segmentation,'' \emph{IEEE Transactions on Circuits
  and Systems for Video Technology}, 2023.

\bibitem{aaunet}
G.~Chen, L.~Li, Y.~Dai, J.~Zhang, and M.~H. Yap, ``Aau-net: an adaptive
  attention u-net for breast lesions segmentation in ultrasound images,''
  \emph{IEEE Transactions on Medical Imaging}, vol.~42, no.~5, pp. 1289--1300,
  2022.

\bibitem{h2former}
A.~He, K.~Wang, T.~Li, C.~Du, S.~Xia, and H.~Fu, ``H2former: An efficient
  hierarchical hybrid transformer for medical image segmentation,'' \emph{IEEE
  Transactions on Medical Imaging}, vol.~42, no.~9, pp. 2763--2775, 2023.

\bibitem{cmunet}
F.~Tang, L.~Wang, C.~Ning, M.~Xian, and J.~Ding, ``Cmu-net: a strong
  convmixer-based medical ultrasound image segmentation network,'' in
  \emph{2023 IEEE 20th International Symposium on Biomedical Imaging
  (ISBI)}.\hskip 1em plus 0.5em minus 0.4em\relax IEEE, 2023, pp. 1--5.

\bibitem{csvt1}
L.~Ma, G.~Tan, H.~Luo, Q.~Liao, S.~Li, and K.~Li, ``A novel deep learning
  framework for automatic recognition of thyroid gland and tissues of neck in
  ultrasound image,'' \emph{IEEE Transactions on Circuits and Systems for Video
  Technology}, vol.~32, no.~9, pp. 6113--6124, 2022.

\bibitem{csvt5}
X.~Zhao, Z.~Li, X.~Luo, P.~Li, P.~Huang, J.~Zhu, Y.~Liu, J.~Zhu, M.~Yang,
  S.~Chang \emph{et~al.}, ``Ultrasound nodule segmentation using asymmetric
  learning with simple clinical annotation,'' \emph{IEEE Transactions on
  Circuits and Systems for Video Technology}, 2024.

\bibitem{bianjie1}
R.~N. Czerwinski, D.~L. Jones, and W.~D. O'Brien, ``Detection of lines and
  boundaries in speckle images-application to medical ultrasound,'' \emph{IEEE
  Transactions on Medical Imaging}, vol.~18, no.~2, pp. 126--136, 1999.

\bibitem{noise1}
R.~Rosa and F.~C. Monteiro, ``Performance analysis of speckle ultrasound image
  filtering,'' \emph{Computer Methods in Biomechanics and Biomedical
  Engineering: Imaging \& Visualization}, vol.~4, no. 3-4, pp. 193--201, 2016.

\bibitem{banzidong1}
T.~Sakinis, F.~Milletari, H.~Roth, P.~Korfiatis, P.~Kostandy, K.~Philbrick,
  Z.~Akkus, Z.~Xu, D.~Xu, and B.~J. Erickson, ``Interactive segmentation of
  medical images through fully convolutional neural networks,'' \emph{arXiv
  preprint arXiv:1903.08205}, 2019.

\bibitem{banzidong3}
J.~Sun, Y.~Shi, Y.~Gao, L.~Wang, L.~Zhou, W.~Yang, and D.~Shen, ``Interactive
  medical image segmentation via point-based interaction and sequential patch
  learning,'' \emph{arXiv preprint arXiv:1804.10481}, 2018.

\bibitem{banzidong2}
G.~Wang, M.~A. Zuluaga, W.~Li, R.~Pratt, P.~A. Patel, M.~Aertsen, T.~Doel,
  A.~L. David, J.~Deprest, S.~Ourselin \emph{et~al.}, ``Deepigeos: a deep
  interactive geodesic framework for medical image segmentation,'' \emph{IEEE
  Transactions on Pattern Analysis and Machine Intelligence}, vol.~41, no.~7,
  pp. 1559--1572, 2018.

\bibitem{masam}
C.~Chen, J.~Miao, D.~Wu, Z.~Yan, S.~Kim, J.~Hu, A.~Zhong, Z.~Liu, L.~Sun, X.~Li
  \emph{et~al.}, ``Ma-sam: Modality-agnostic sam adaptation for 3d medical
  image segmentation,'' \emph{arXiv preprint arXiv:2309.08842}, 2023.

\bibitem{sammed2d}
J.~Cheng, J.~Ye, Z.~Deng, J.~Chen, T.~Li, H.~Wang, Y.~Su, Z.~Huang, J.~Chen,
  L.~Jiang \emph{et~al.}, ``Sam-med2d,'' \emph{arXiv preprint
  arXiv:2308.16184}, 2023.

\bibitem{samu}
G.~Deng, K.~Zou, K.~Ren, M.~Wang, X.~Yuan, S.~Ying, and H.~Fu, ``Sam-u:
  Multi-box prompts triggered uncertainty estimation for reliable sam in
  medical image,'' in \emph{International Conference on Medical Image Computing
  and Computer-Assisted Intervention}.\hskip 1em plus 0.5em minus 0.4em\relax
  Springer, 2023, pp. 368--377.

\bibitem{desam}
Y.~Gao, W.~Xia, D.~Hu, and X.~Gao, ``Desam: Decoupling segment anything model
  for generalizable medical image segmentation,'' \emph{arXiv preprint
  arXiv:2306.00499}, 2023.

\bibitem{medsam}
J.~Ma, Y.~He, F.~Li, L.~Han, C.~You, and B.~Wang, ``Segment anything in medical
  images,'' \emph{Nature Communications}, vol.~15, no.~1, p. 654, 2024.

\bibitem{msa}
J.~Wu, W.~Ji, Y.~Liu, H.~Fu, M.~Xu, Y.~Xu, and Y.~Jin, ``Medical sam adapter:
  Adapting segment anything model for medical image segmentation,'' \emph{arXiv
  preprint arXiv:2304.12620}, 2023.

\bibitem{samed}
K.~Zhang and D.~Liu, ``Customized segment anything model for medical image
  segmentation,'' \emph{arXiv preprint arXiv:2304.13785}, 2023.

\bibitem{evidentialdeeplearining}
M.~Sensoy, L.~Kaplan, and M.~Kandemir, ``Evidential deep learning to quantify
  classification uncertainty,'' \emph{Advances in Neural Information Processing
  Systems}, vol.~31, 2018.

\bibitem{dirichletuncertainty}
M.~Xie, S.~Li, R.~Zhang, and C.~H. Liu, ``Dirichlet-based uncertainty
  calibration for active domain adaptation,'' \emph{arXiv preprint
  arXiv:2302.13824}, 2023.

\bibitem{graphcut}
Y.~Y. Boykov and M.-P. Jolly, ``Interactive graph cuts for optimal boundary \&
  region segmentation of objects in nd images,'' in \emph{Proceedings Eighth
  IEEE International Conference on Computer Vision. ICCV 2001}, vol.~1.\hskip
  1em plus 0.5em minus 0.4em\relax IEEE, 2001, pp. 105--112.

\bibitem{randomwalks}
L.~Grady, ``Random walks for image segmentation,'' \emph{IEEE Transactions on
  Pattern Analysis and Machine Intelligence}, vol.~28, no.~11, pp. 1768--1783,
  2006.

\bibitem{geos}
V.~Gulshan, C.~Rother, A.~Criminisi, A.~Blake, and A.~Zisserman, ``Geodesic
  star convexity for interactive image segmentation,'' in \emph{2010 IEEE
  Computer Society Conference on Computer Vision and Pattern
  Recognition}.\hskip 1em plus 0.5em minus 0.4em\relax IEEE, 2010, pp.
  3129--3136.

\bibitem{grabcut}
C.~Rother, V.~Kolmogorov, and A.~Blake, ``" grabcut" interactive foreground
  extraction using iterated graph cuts,'' \emph{ACM Transactions on Graphics
  (TOG)}, vol.~23, no.~3, pp. 309--314, 2004.

\bibitem{gailv6}
C.~F. Baumgartner, K.~C. Tezcan, K.~Chaitanya, A.~M. H{\"o}tker, U.~J.
  Muehlematter, K.~Schawkat, A.~S. Becker, O.~Donati, and E.~Konukoglu,
  ``Phiseg: Capturing uncertainty in medical image segmentation,'' in
  \emph{Medical Image Computing and Computer Assisted Intervention--MICCAI
  2019: 22nd International Conference, Shenzhen, China, October 13--17, 2019,
  Proceedings, Part II 22}.\hskip 1em plus 0.5em minus 0.4em\relax Springer,
  2019, pp. 119--127.

\bibitem{gailv4}
L.~Hu, J.~Li, X.~Peng, J.~Xiao, B.~Zhan, C.~Zu, X.~Wu, J.~Zhou, and Y.~Wang,
  ``Semi-supervised npc segmentation with uncertainty and attention guided
  consistency,'' \emph{Knowledge-Based Systems}, vol. 239, p. 108021, 2022.

\bibitem{gailv5}
S.~Kohl, B.~Romera-Paredes, C.~Meyer, J.~De~Fauw, J.~R. Ledsam, K.~Maier-Hein,
  S.~Eslami, D.~Jimenez~Rezende, and O.~Ronneberger, ``A probabilistic u-net
  for segmentation of ambiguous images,'' \emph{Advances in Neural Information
  Processing Systems}, vol.~31, 2018.

\bibitem{gailv1}
T.~Nair, D.~Precup, D.~L. Arnold, and T.~Arbel, ``Exploring uncertainty
  measures in deep networks for multiple sclerosis lesion detection and
  segmentation,'' \emph{Medical Image Analysis}, vol.~59, p. 101557, 2020.

\bibitem{gailv3}
P.~Tang, P.~Yang, D.~Nie, X.~Wu, J.~Zhou, and Y.~Wang, ``Unified medical image
  segmentation by learning from uncertainty in an end-to-end manner,''
  \emph{Knowledge-Based Systems}, vol. 241, p. 108215, 2022.

\bibitem{gailv2}
G.~Wang, W.~Li, M.~Aertsen, J.~Deprest, S.~Ourselin, and T.~Vercauteren,
  ``Aleatoric uncertainty estimation with test-time augmentation for medical
  image segmentation with convolutional neural networks,''
  \emph{Neurocomputing}, vol. 338, pp. 34--45, 2019.

\bibitem{uncercsvt}
T.~Zhou, Y.~Zhou, G.~Li, G.~Chen, and J.~Shen, ``Uncertainty-aware hierarchical
  aggregation network for medical image segmentation,'' \emph{IEEE Transactions
  on Circuits and Systems for Video Technology}, 2024.

\bibitem{uncercsvt2}
B.~Xie, L.~Yuan, S.~Li, C.~H. Liu, and X.~Cheng, ``Towards fewer annotations:
  Active learning via region impurity and prediction uncertainty for domain
  adaptive semantic segmentation,'' in \emph{Proceedings of the IEEE/CVF
  conference on computer vision and pattern recognition}, 2022, pp. 8068--8078.

\bibitem{uncercsvt3}
Z.~Luo, X.~Luo, Z.~Gao, and G.~Wang, ``An uncertainty-guided tiered
  self-training framework for active source-free domain adaptation in prostate
  segmentation,'' in \emph{International Conference on Medical Image Computing
  and Computer-Assisted Intervention}.\hskip 1em plus 0.5em minus 0.4em\relax
  Springer, 2024, pp. 107--117.

\bibitem{jicheng2}
T.~Buddenkotte, L.~E. Sanchez, M.~Crispin-Ortuzar, R.~Woitek, C.~McCague, J.~D.
  Brenton, O.~{\"O}ktem, E.~Sala, and L.~Rundo, ``Calibrating ensembles for
  scalable uncertainty quantification in deep learning-based medical image
  segmentation,'' \emph{Computers in Biology and Medicine}, vol. 163, p.
  107096, 2023.

\bibitem{jicheng4}
B.~Lakshminarayanan, A.~Pritzel, and C.~Blundell, ``Simple and scalable
  predictive uncertainty estimation using deep ensembles,'' \emph{Advances in
  Neural Information Processing Systems}, vol.~30, 2017.

\bibitem{jicheng1}
A.~Mehrtash, W.~M. Wells, C.~M. Tempany, P.~Abolmaesumi, and T.~Kapur,
  ``Confidence calibration and predictive uncertainty estimation for deep
  medical image segmentation,'' \emph{IEEE Transactions on Medical Imaging},
  vol.~39, no.~12, pp. 3868--3878, 2020.

\bibitem{jicheng3}
S.~Rosas-Gonzalez, T.~Birgui-Sekou, M.~Hidane, I.~Zemmoura, and C.~Tauber,
  ``Asymmetric ensemble of asymmetric u-net models for brain tumor segmentation
  with uncertainty estimation,'' \emph{Frontiers in Neurology}, vol.~12, p.
  609646, 2021.

\bibitem{evi4}
Y.~Chen, Z.~Yang, C.~Shen, Z.~Wang, Z.~Zhang, Y.~Qin, X.~Wei, J.~Lu, Y.~Liu,
  and Y.~Zhang, ``Evidence-based uncertainty-aware semi-supervised medical
  image segmentation,'' \emph{Computers in Biology and Medicine}, vol. 170, p.
  108004, 2024.

\bibitem{evi1}
Y.~He, ``Epl: Evidential prototype learning for semi-supervised medical image
  segmentation,'' \emph{arXiv preprint arXiv:2404.06181}, 2024.

\bibitem{evi3}
Y.~Yang, X.~Xu, H.~Hu, H.~Long, Q.~Zhou, and Q.~Guan, ``Duedl: Dual-branch
  evidential deep learning for scribble-supervised medical image
  segmentation,'' \emph{arXiv preprint arXiv:2405.14444}, 2024.

\bibitem{evi2}
Z.~Zhang, H.~Zhou, X.~Shi, R.~Ran, C.~Tian, and F.~Zhou, ``An
  evidential-enhanced tri-branch consistency learning method for
  semi-supervised medical image segmentation,'' \emph{arXiv preprint
  arXiv:2404.07032}, 2024.

\bibitem{evi5}
K.~Zou, X.~Yuan, X.~Shen, M.~Wang, and H.~Fu, ``Tbrats: Trusted brain tumor
  segmentation,'' in \emph{International Conference on Medical Image Computing
  and Computer-Assisted Intervention}.\hskip 1em plus 0.5em minus 0.4em\relax
  Springer, 2022, pp. 503--513.

\bibitem{DST}
A.~P. Dempster, ``A generalization of bayesian inference,'' \emph{Journal of
  the Royal Statistical Society: Series B (Methodological)}, vol.~30, no.~2,
  pp. 205--232, 1968.

\bibitem{sl}
A.~Jsang, \emph{Subjective Logic: A formalism for reasoning under
  uncertainty}.\hskip 1em plus 0.5em minus 0.4em\relax Springer Publishing
  Company, Incorporated, 2018.

\bibitem{busi}
W.~Al-Dhabyani, M.~Gomaa, H.~Khaled, and A.~Fahmy, ``Dataset of breast
  ultrasound images,'' \emph{Data in brief}, vol.~28, p. 104863, 2020.

\bibitem{ddti}
L.~Pedraza, C.~Vargas, F.~Narv{\'a}ez, O.~Dur{\'a}n, E.~Mu{\~n}oz, and
  E.~Romero, ``An open access thyroid ultrasound image database,'' in
  \emph{10th International Symposium on Medical Information Processing and
  Analysis}, vol. 9287.\hskip 1em plus 0.5em minus 0.4em\relax SPIE, 2015, pp.
  188--193.

\bibitem{echo}
D.~Ouyang, B.~He, A.~Ghorbani, N.~Yuan, J.~Ebinger, C.~P. Langlotz, P.~A.
  Heidenreich, R.~A. Harrington, D.~H. Liang, E.~A. Ashley \emph{et~al.},
  ``Video-based ai for beat-to-beat assessment of cardiac function,''
  \emph{Nature}, vol. 580, no. 7802, pp. 252--256, 2020.

\bibitem{paszke2019pytorch}
A.~Paszke, S.~Gross \emph{et~al.}, ``Pytorch: An imperative style,
  high-performance deep learning library,'' \emph{Advances in Neural
  Information Processing Systems}, vol.~32, 2019.

\bibitem{segnet}
V.~Badrinarayanan, A.~Kendall, and R.~Cipolla, ``Segnet: A deep convolutional
  encoder-decoder architecture for image segmentation,'' \emph{IEEE
  Transactions on Pattern Analysis and Machine Intelligence}, vol.~39, no.~12,
  pp. 2481--2495, 2017.

\bibitem{transunet}
J.~Chen, Y.~Lu, Q.~Yu, X.~Luo, E.~Adeli, Y.~Wang, L.~Lu, A.~L. Yuille, and
  Y.~Zhou, ``Transunet: Transformers make strong encoders for medical image
  segmentation,'' \emph{arXiv preprint arXiv:2102.04306}, 2021.

\bibitem{sam}
A.~Kirillov, E.~Mintun, N.~Ravi, H.~Mao, C.~Rolland, L.~Gustafson, T.~Xiao,
  S.~Whitehead, A.~C. Berg, W.-Y. Lo \emph{et~al.}, ``Segment anything,'' in
  \emph{Proceedings of the IEEE/CVF International Conference on Computer
  Vision}, 2023, pp. 4015--4026.

\end{thebibliography}

\end{document}